\pdfoutput=1

\documentclass[11pt]{article}

\usepackage{acl}

\usepackage{times}
\usepackage{latexsym}
\usepackage{comment}
\usepackage{subcaption}

\usepackage[T1]{fontenc}

\usepackage[utf8]{inputenc}

\usepackage{microtype}

\usepackage{tabularx}

\usepackage{booktabs}
\usepackage{xcolor}
\usepackage{soul}
\usepackage{graphicx}
\usepackage{amsmath}

\newcommand{\fref}[1]{Figure~\ref{#1}}
\newcommand{\tref}[1]{Table~\ref{#1}}

\newcommand*\samethanks[1][\value{footnote}]{\footnotemark[#1]}

\title{When ``A Helpful Assistant'' Is Not Really Helpful: Personas in System Prompts Do Not Improve  Performances of Large Language Models}

\author{
    Mingqian Zheng $^{\heartsuit}$\thanks{~~Work done as students at the University of Michigan} ~ Jiaxin Pei $^{\dagger}$\samethanks \ ~ Lajanugen Logeswaran $^{\sharp}$ \\
    ~ \textbf{Moontae Lee} $^{\sharp \diamond}$ ~ \textbf{David Jurgens} $^{\ddagger}$ \vspace{0.1cm}\\
    $^\heartsuit$ Carnegie Mellon University ~
    $^{\dagger}$ Stanford University \vspace{0.1cm} \\
    $^{\sharp}$ LG AI Research ~
    $^{\diamond}$ University of Illinois Chicago ~
    $^{\ddagger}$ University of Michigan \\
    { \tt $^{\heartsuit}$mingqia2@andrew.cmu.edu} ~
    { \tt $^{\dagger}$pedropei@stanford.edu}\\
    { \tt $^{\sharp}$\{llajan, moontae.lee\}@lgresearch.ai} 
    { \tt $^{\ddagger}$jurgens@umich.edu} \\
}

\begin{document}
\maketitle
\begin{abstract}
Prompting serves as the major way humans interact with Large Language Models (LLM). Commercial AI systems commonly define the role of the LLM in system prompts. For example, ChatGPT uses ``You are a helpful assistant'' as part of its default system prompt. Despite current practices of adding personas to system prompts, it remains unclear how different personas affect a model's performance on objective tasks. In this study, we present a systematic evaluation of personas in system prompts. We curate a list of 162 roles covering 6 types of interpersonal relationships and 8 domains of expertise. Through extensive analysis of 4 popular families of LLMs and 2,410 factual questions, we demonstrate that adding personas in system prompts does not improve model performance across a range of questions compared to the control setting where no persona is added. Nevertheless, further analysis suggests that the gender, type, and domain of the persona can all influence the resulting prediction accuracies. We further experimented with a list of persona search strategies and found that, while aggregating results from the best persona for each question significantly improves prediction accuracy, automatically identifying the best persona is challenging, with predictions often performing no better than random selection. Overall, our findings suggest that while adding a persona may lead to performance gains in certain settings, the effect of each persona can be largely random. Code and data are available at \url{https://github.com/Jiaxin-Pei/Prompting-with-Social-Roles}.
\end{abstract}

\section{Introduction}

Building persona- or role-based chatbots has attracted enormous attention from the AI and NLP community due to their values for potential business and societal applications \citep{pataranutaporn2021ai}. Recent advances in LLMs provide huge opportunities to build intelligent agents that can behave and talk like certain characters or roles \citep{wang2023rolellm}. Despite all the existing studies on LLM role-playing, it is unclear how different personas affect LLMs' performance on objective tasks. To address this gap, we conduct a large-scale analysis of 162 personas over 4 popular families of open-source LLMs and 2410 factual questions. To ensure the generalizability of the result, the 162 personas are selected from 6 types of interpersonal relationships and 8 domains of expertise. Furthermore, to study the effect of domain alignment between personas and questions, the evaluation question sets are sampled from the Massive Multitask Language Understanding (MMLU) dataset \citep{hendryckstest2021}, balanced for categories.

\begin{figure}[t!]
    \centering    \includegraphics[width=\columnwidth]{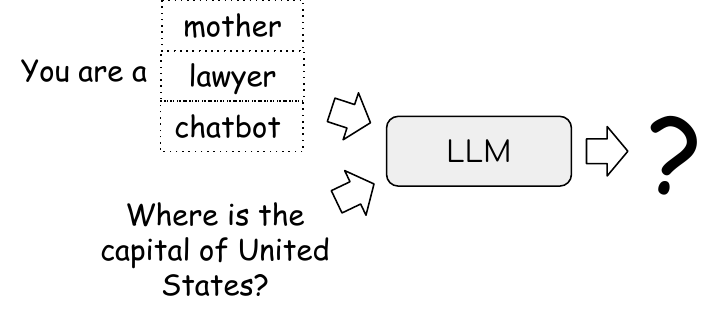}
    \caption{Our overall research question: does adding personas in prompts affect LLMs' performance?}
    \label{fig:illustration}
\end{figure}

In this study, we aim to answer four major research questions: (1) Does adding personas to system prompts help improve model performance on objective tasks? (2) Does the social construct of the persona affect model performance? (3) What factors could potentially explain the effect of personas on model performance? (4) Can we automatically identify the best roles for prompting? Through our analysis, we find that, in general, prompting with personas has no or small negative effects on model performance compared with the control setting where no persona is added. This result is consistent across four popular LLM families, suggesting that adding personas into system prompts may not help improve the model's performance. To further understand the relative differences among personas, we analyze the social attributes of personas, including role type, gender, and domain alignment. We find that gender-neutral, in-domain, and work-related roles lead to better performance than other types of roles, but with relatively small effect sizes, suggesting that the social construct of the persona may not fully explain the consequential performance differences.

To understand the potential mechanisms behind the relative performance differences caused by different personas, we further analyze the word frequency of the personas, the similarity between prompt-questions pairs, and the perplexity. We observe that personas with high-frequency words lead to relatively better model performance. Furthermore, while the similarity between the persona and the question is the strongest predictor of final performance, the correlation between prompt-question similarity and prediction accuracy remains low. Our results suggest that word frequency, prompt-question similarity, and perplexity do not explain much of the performance gaps between different personas. To further uncover the potential mechanism behind persona-based prompting, we try to approach this from another perspective: automatic persona search. The hypothesis is that if the effect of persona varies in a systematic way, we should be able to identify which persona might result in better performance automatically.  We test a list of automatic persona selection strategies, from simple random selection to fine-tuning RoBERTa-based classifiers. The effect of each persona on model performance varies across questions, making it difficult to reliably identify which persona consistently yields better performance. This suggests that the effect of personas on LLMs' performance might largely be random. 

Our study makes the following three contributions. First, we introduce a new pipeline to systematically evaluate LLMs' performance when prompted with a wide range of personas. Second, our large-scale experiments reveal an important finding that prompting LLMs with personas might actually hurt their performance on objective tasks. Third,  through analyzing a wide range of persona attributes and automatic role-searching strategies, we found that the effect of personas on model performance is not consistent across questions. While a certain persona may lead to the correct answer for each question, the presence of such personas is largely unpredictable.

\section{Related work}

\paragraph{Personas and Roles}
Personas are fundamental in human society and day-to-day interactions \citep{heiss2017social, goffman2016presentation}.  personas define the norm of human interactions and affect human behaviors in various contexts \citep{sunstein1996social}. Two prominent types of personas are interpersonal roles which are deeply embedded in interpersonal relationships \citep{berscheid1994interpersonal} (e.g., mother and friend), and professional/occupational roles that fulfill certain social functions or provide certain services in the society (e.g., driver and teacher) \citep{bucher1961professions, brante1988sociological}.
As suggested by \citet{wolfensberger2000brief}, ``People largely perceive themselves and each other in terms of their roles.'' Given the importance of personas in human interactions and recent advances in persona-based agents \citep{wang2023rolellm, pataranutaporn2021ai, gupta2023bias}, understanding LLMs' role-playing capabilities and the effect of personas hold significance to both the NLP community and the general public. Previous literature examines the influence of in-context impersonation in vision-based reasoning tasks \cite{salewski2024context} and shows that providing sociodemographic information to models may benefit subjective NLP tasks in zero-shot settings. However, some other studies also point out the potential bias and limitations of such persona and social-demographic-based prompting \citep{sun2023aligning, hu2024quantifying, beck2024sensitivity}. 
While existing studies on subjective tasks uncover important patterns in models' performance changes when prompted with personas, teasing out the effect of personas on models' performances is hard because of the natural subjectivity of the task. Therefore, our study focuses on objective tasks where the models' performance change is solely affected by the added persona.

\paragraph{Prompting LLM}
Prompting serves as a unified natural language interface for human-AI interactions and has been widely adopted in the era of LLM \citep{liu2023pre}. Existing studies suggest that LLMs are very sensitive to the design of prompts \citep{lu2021fantastically}. For example, adding ``Let's think step by step'' could help to improve the model performance in answering a wide range of questions \citep{kojima2022large}. How to design prompts that lead to better performance has become an important question for not only NLP researchers but also people in education \citep{heston2023prompt}, art \citep{oppenlaender2022prompt} and health \cite{mesko2023prompt} industries. Furthermore, current AI systems usually insert system prompts before user prompts to ensure the safety and helpfulness of system-generated outputs \citep{touvron2023llama}. System prompts usually define the role of the system (e.g. ``You are a helpful assistant.'') and further guide LLMs' behaviors in user interactions. That is, the system prompt serves as a default setting of LLM products and precedes any user prompt. Thus, even for models that are not instruction-tuned, it is still important to investigate how various formatted system prompts might impact model performance. Despite its wide usage in commercial AI systems, the effect of using personas in systems prompts has not been fully studied in the current literature. 

\paragraph{Role Playing with LLMs}
Creating agents that are able to talk like certain characters and roles has attracted much attention from the AI and NLP community \citep{demasi2020multi} due to its potential benefits in settings like education \citep{pataranutaporn2021ai}, games \citep{miikkulainen2007creating}, and mental health \citep{denecke2020mental}. Large language models offer new opportunities in creating persona-based agents through role-playing with LLMs \citep{shanahan2023role}. Existing studies have produced datasets \citep{qian2021pchatbot}, prompting strategies \citep{kong2023better}, and evaluation settings \citep{wang2023rolellm} for role-playing with LLMs. However, when evaluating LLMs' role-playing capabilities, existing studies majorly focus on role- and dialogue-related metrics such as perplexity, coherence, and interestingness \citep{lin2020xpersona, deriu2021survey}. Prompting models to role-play may lead to negative social effects \cite{gupta2023bias} and needs to be evaluated comprehensively \cite{cheng2023compost}. It is still unclear whether role-playing would affect LLMs' capability to handle general language tasks. 

\section{Experiment Setting}
Our study aims to test whether adding personas in prompts affects LLMs' performances. In this section, we describe the experiment setup to examine the persona effect on models' performances.

\subsection{Dataset}
We use a subset of MMLU \cite{hendryckstest2021} for all of our experiments. MMLU, a dataset designed for multitask language understanding, is widely used to benchmark LLMs. It features multiple-choice questions that probe knowledge across a diverse set of subjects, ranging from natural sciences and social sciences to business and law. We choose MMLU as our test dataset because (1) it has been widely used for benchmarking LLMs, (2) it includes questions from diverse disciplines, enabling us to test the effect of prompting with domain-aligned personas, and (3) questions across different domains follow similar formats, which reduces potential confounds.

Furthermore, to ensure the generalizability of our results, we design a sampling pipeline to balance the length and subject of the question.  We first randomly sample 100 instances from each initial subject of MMLU to ensure a diverse representation of questions across subjects. For each sampled instance, we calculate the length of full questions with both question text and four options. To manage the computation cost, we drop questions so that 99\% of the sampled questions have fewer than 150 words. From the filtered dataset, we manually select subjects based on higher popularity and coverage of several broad domains. The final dataset contains 2410 questions from the MMLU dataset, balanced across 26 subjects. We further map the sampled subjects into eight core categories: Law, Medicine, Computer Science, Math, Politics, Psychology, Natural Science, and Economics. \tref{tab:domain_mapping} in the Appendix details the subjects and domains. 

\begin{table}[t]
    \centering
    \resizebox{0.48\textwidth}{!}{
    \begin{tabular}{ll}
        \textbf{Prompt Type} & \textbf{Example}  \\
    \hline
        No Role & \{question\}  \\
        Speaker-Specific & You are a/an \{role\}, \{question\}\\
        Audience-Specific & You are talking to a/an \{role\}, \{question\} 
    \end{tabular}}
    \caption{Types and examples of prompt templates for personas used in our experiment. We further refine the prompt to meet the format requirement of each model, and the full prompts are available in the Appendix (\tref{tab:context_prompts} and \tref{tab:control_prompts}).}
    \label{tab:prompt}
\end{table}

\subsection{Prompt}
Personas can be incorporated into prompts in various ways. We carefully design two types of prompts: (1) \textbf{Speaker-Specific Prompt:} prompts that assign the role to the LLM (i.e., ``who you are''). For example, ``You are a lawyer''; (2) \textbf{Audience-Specific Prompt:} prompts that specify the audience of the conversation (i.e., ``whom you are talking to''). For example, ``You are talking to a fireman''. As a comparison, prompts that only include the question are used as the control setting in our experiment. Table ~\ref{tab:prompt} shows the template of prompts used in our study. As a robustness check, for each prompt template, we also include an external paraphrased prompt by adding the word ``Imagine'' (e.g. ``Imagine you are talking to a fireman''). We further revise the prompt template to fit into the format requirements of different models to attain the best performance. Tables~\ref{tab:context_prompts} and \ref{tab:control_prompts} in the Appendix detail the prompt we use for each model.

\subsection{Persona}
To excessively evaluate the effect of personas on model performance, we curate a large and diverse list of personas that are actively used in people's daily interactions.  We first collect over 300 personas based on several existing studies \citep{garg2018word, massey2015annotating, choi2021more}, WordNet \citep{miller1995wordnet}, and our own ad-hoc social role list. We manually examine the roles to remove uncommon roles that are rarely used in daily life, such as ``ganger'' as a hyponym for ``boss''. Our final social role set includes 162 personas, of which 112 roles are occupations, and the remaining are interpersonal relationship roles. \tref{tab:all_roles} in the Appendix shows the full list of roles in our experiment.

\paragraph{Interpersonal Roles}
Our study includes 50 interpersonal roles grouped into 5 categories: family, friend, romantic, work, and school. For important roles that do not fit into the above categories (e.g. stranger), we add them into the category of ``social''. 
We further augment the role list by adding hyponyms from WordNet \citep{miller1995wordnet} to selected roles as a robustness check. 
For example, for the word ``mother'', we also include ``mama'', ``mamma'', ``mom'' and ``mommy''. 

\paragraph{Occupational Roles}
We compile our set of occupations from \citet{garg2018word}. Additionally, we manually add occupations that are relevant to the subjects of the sampled MMLU questions. For example, we add ``software engineer'' under the category of Computer Science. Furthermore, given the wide adoption of AI systems in our society, we also include a list of AI roles (e.g. ``AI language model'' and ``AI assistant'').

\subsection{Models}

We experiment with 9 popular open-source instruction-tuned LLM from 4 model collections: FLAN-T5-XXL (11B) \cite{flant5}, Llama-3-Instruct (8B and 70B) \cite{llama3modelcard}, Mistral-7B-Instruct-v0.2 \cite{jiang2023mistral} and Qwen2.5-Instruct (3B to 72B) \cite{qwen2.5}. Our main experiments are conducted on one midsized of $\sim$7B and one large-sized model of $\sim$70B when available for each model collection. We also later use multiple intermediate-sized models from the Qwen2.5 collection to analyze scaling effects. 

All of the four model collections have already been fine-tuned to follow instructions, and all of them except for Flan-T5 allow a chat template that contains both a system prompt and a user prompt. We choose open-weight models mainly because of the following reasons: (1) 7 to 11B open-weight models have shown promising performances on a wide range of tasks, especially Llama-3 and Qwen. Smaller-size models may not have enough role-playing or instruction-following capabilities; (2) our experiment requires running inference tasks over 2410 questions with 4 prompt templates and 162 personas, making it computationally and financially expensive to query API-based or bigger models; (3) experimenting with open-weight models allows other researchers to easily replicate our experiment results; (4) models of varying sizes from the same collection allow for the study of the scaling effects of personas. 

\section{Does Prompting with Personas Improve LLMs' Performance?}

\begin{figure}[t]
    \centering    
    \includegraphics[width=\columnwidth]{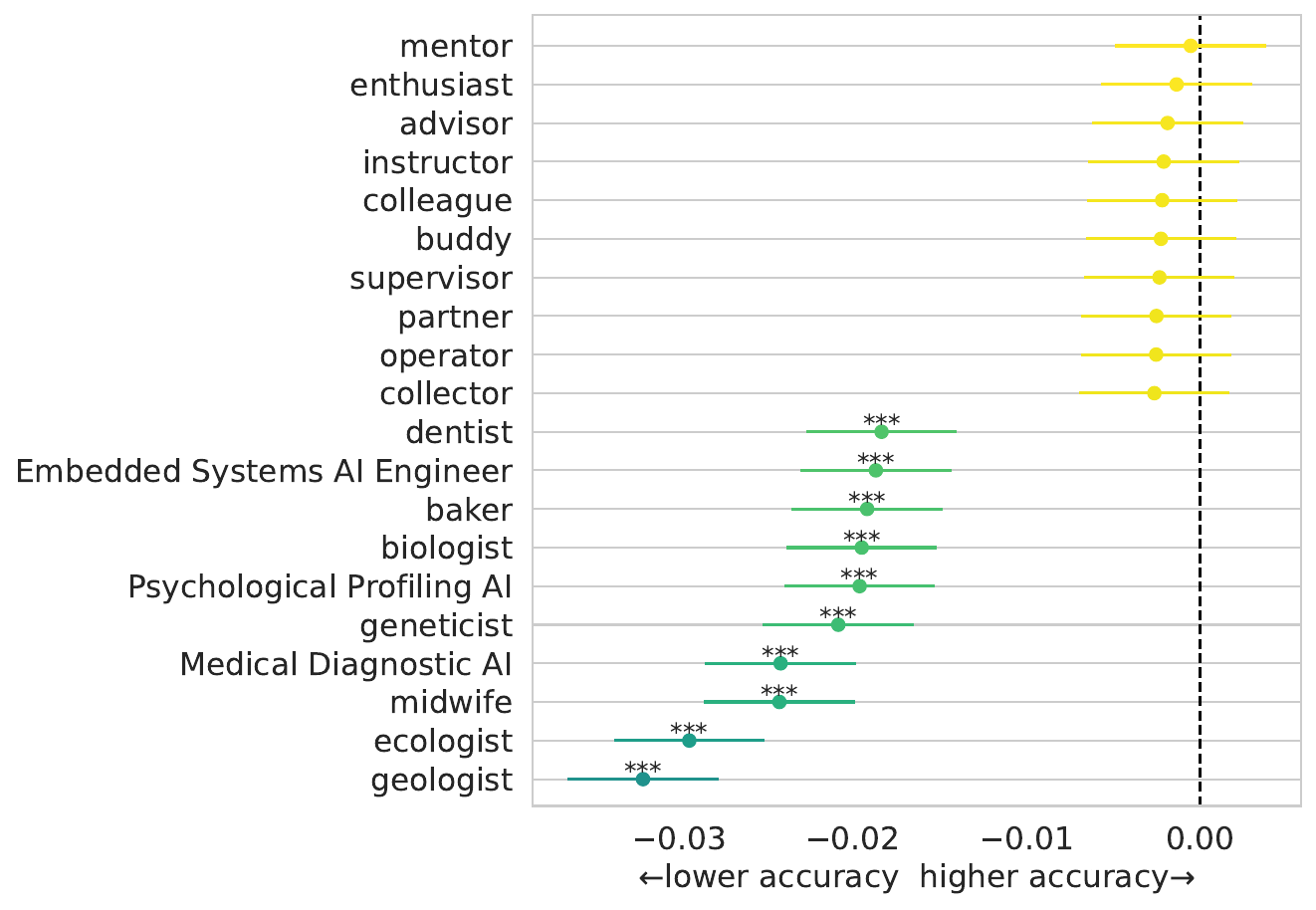}
    \caption{The first and last 10 personas ranked by their effect sizes on the models' performance change compared with the control setting. Overall, none of the personas lead to statistically better model performance.}
    \label{fig:regression-role-mlm-top10}
\end{figure}

To assess whether adding personas helps improve model performance in answering factual questions, we fit a mixed-effects regression model that uses the added persona to predict the inference accuracy, controlling a random effect for each model to account for potential variability across different models. The control setting, where no role is added to the system prompts, is used as the reference category. Figure~\ref{fig:regression-role-mlm-top10} shows the first and last 10 coefficients ranked by their effect sizes on the models' performance change compared with the control setting. The coefficients for all roles are detailed in Section \ref{appendix:role-regression} in the Appendix. We observe no significant differences between the best-performing personas and the control setting. Additionally, linear regression results for each model are also listed in Appendix~\ref{appendix:role-regression}.

\begin{figure}[t]
    \centering    
    \includegraphics[width=\columnwidth]{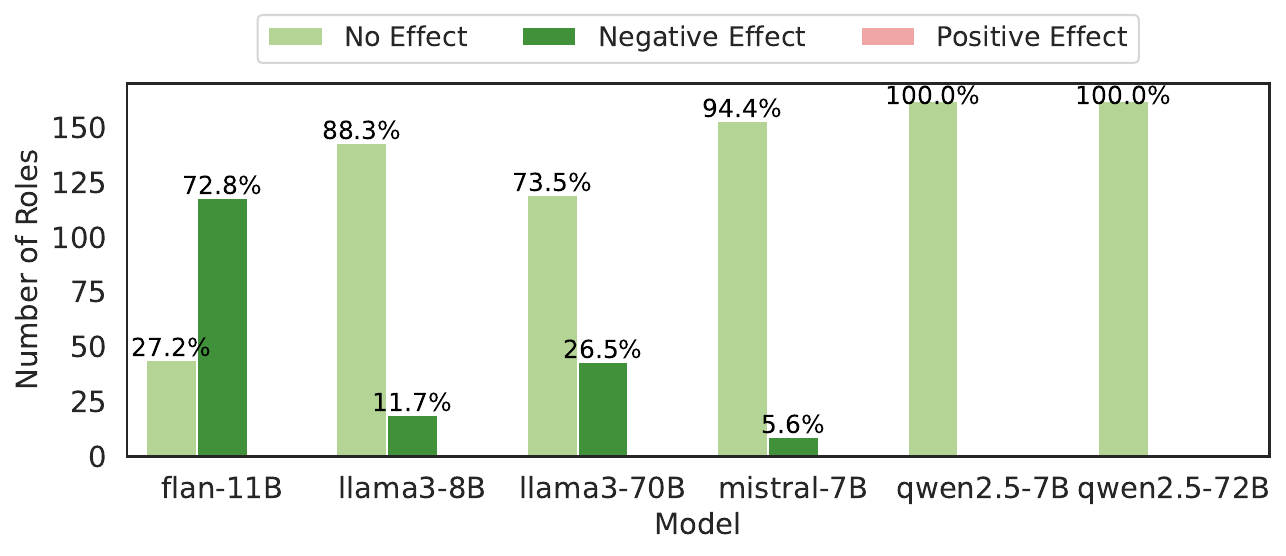}
    \caption{Most of the personas have no or negative impact on LLM's performance.}
    \label{fig:regression-role-type}
\end{figure}

\begin{figure}[t]
    \centering    
    \includegraphics[width=\columnwidth]{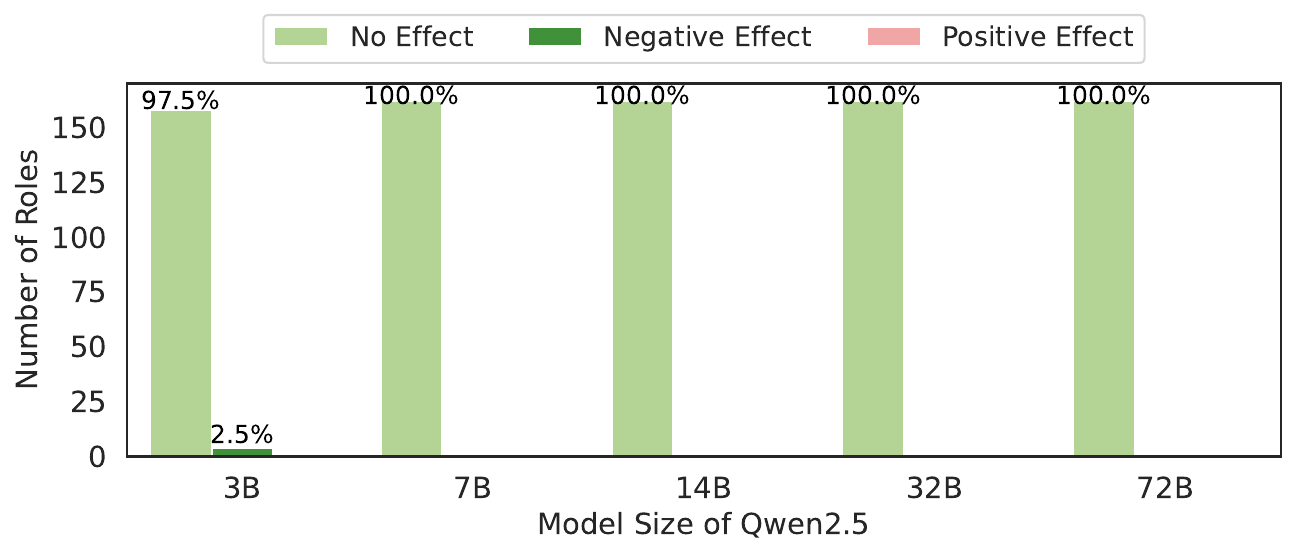}
    \caption{LLMs' sensitivity to persona does not scale with model sizes}
    \label{fig:regression-bar-qwen}
\end{figure}

On the contrary, certain personas may actually lead to \textit{lower} performance (e.g., ecologist for Mistral). As shown in Figure~\ref{fig:regression-role-type}, most of the personas have no statistically significant effect on the model's prediction accuracy compared with the control setting, and such a pattern is consistent across all six models. Considering the model size, we observe that for the larger model Llama3-70B, more personas have negative effects, indicating a potential scaling effect, while Qwen2.5-7B and Qwen2.5-72B are insensitive to all 162 personas. Furthermore, Figure~\ref{fig:regression-bar-qwen} demonstrates that persona effects are insensitive to model sizes. Our results suggest that there might not exist a single persona that can consistently help to improve LLMs' performance across diverse questions. 

\begin{figure}[t]
    \centering    
    \includegraphics[width=\columnwidth]{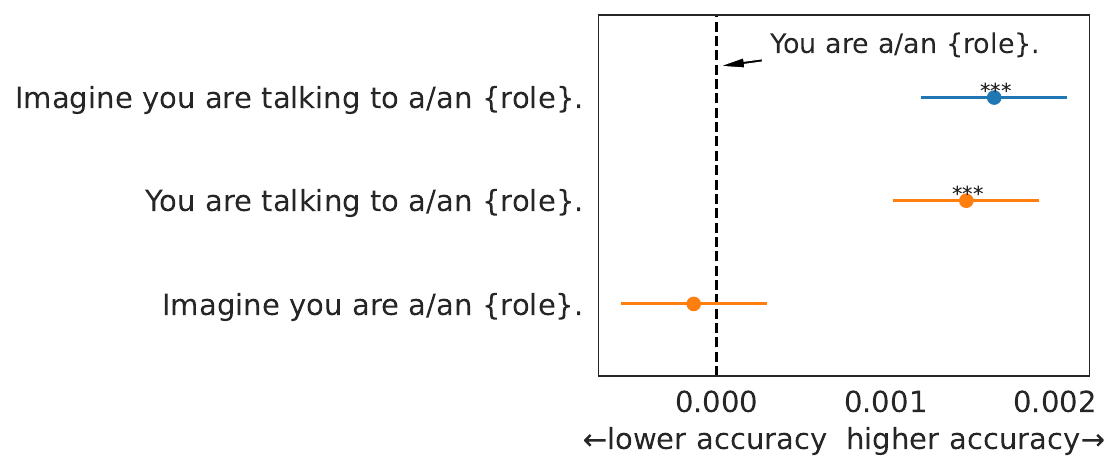}
    \caption{Audience-specific prompts are significantly better than speaker-specific prompts with small effect sizes.}
    \label{fig:acc-prompt-mlm2}
\end{figure}

Does the framing of the prompt affect the model's performance? To answer this question, we run a mixed-effects model on the relationship between accuracy and prompt type, controlling for each model as a random effect. Figure~\ref{fig:acc-prompt-mlm2} shows the regression coefficients for each prompt template. We observe that audience-specific prompts perform better than speaker-specific prompts, and the difference is statistically significant. However, we must note that the effect size is relatively small, suggesting that different framings of the prompt have limited impacts on model performance.

\section{Are Certain Personas Better Than Others?}
While adding a persona might not outperform the control setting with no role, in practice, LLM service providers or users may still need to define the role of the system for various reasons (e.g., security and language styles). Therefore, it is still worth testing whether different categories of personas could lead to systematically different performances.   

\paragraph{Gender} 
Gender roles are one of the most prominent and widely studied personas in Sociology \citep{blackstone2003gender, acker1992sex}. In language, gender is explicitly marked in various types of personas like father and wife; however, gender may be implicit and inferred for some roles such as plumber or nurse where the workforce participation rates skew towards one gender.\footnote{In this setting, gender is inferred by the reader (potentially incorrectly), and word representations have been known to contain such biases \citep[e.g.,][]{caliskan2017semantics}.  } Do LLMs exhibit a tendency whereby a ``father'' role is more likely to yield accurate responses compared to a ``mother'' role? To quantify the impact of gender, we assess interpersonal roles and occupational roles separately by analyzing the explicit and implicit gender impact, respectively. 

\begin{figure}[t]
    \centering    
    \includegraphics[width=\columnwidth]{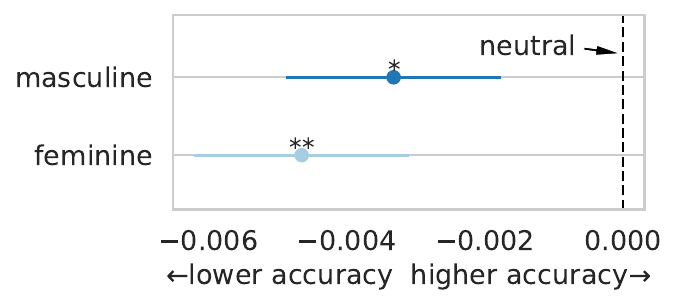}
    \caption{Gender-neutral roles lead to better performances than gendered roles.}
    \label{fig:lme-gender-aligned}
\end{figure}

For interpersonal roles, we analyze 16 aligned roles and categorize them as masculine, feminine, or neutral, resulting in seven masculine roles, seven feminine roles, and two gender-neutral roles. \tref{tab:aligned_gender_roles} in the Appendix shows the mapping of gender and roles. Such a setting allows us to control the effects of role types and reveal the nuanced effects of gender. We employ a mixed-effects model to analyze the relationship between accuracy and gender, with ``accuracy'' as the dependent variable, ``gender'' as an independent categorical variable of values ``masculine'', ``feminine'' and ``neutral'', and we include a random effect for each model. As shown in~\fref{fig:lme-gender-aligned}, gender-neutral roles perform significantly better than gendered roles and masculine roles perform slightly better than feminine roles with a small effect size.  

For occupational roles, we use the percentages of workers belonging to each gender in 65 occupational roles, extracted from historical US census data \cite{garg2018word}. We fit a similar mixed-effects model with the percentage of masculine workers as the independent variable and include random intercepts for each model. The coefficient of ``Masculine'', the percentage of masculine workers for each occupation, is -5.79e-4. The associated p-value is 0.561, indicating that the gender percentage is not a significant predictor of model performance. The results of the two mixed-effects models for gender impact collectively lead to the conclusion that the gender nature of personas has a very limited impact on the models' performance in terms of accuracy. 

\paragraph{Role Category} Our experiment includes 162 roles from 7 groups: work, school, social, family, romantic, occupation, and AI. These categories distinguish roles based on the social relationships and settings they typically involve. Does the effect of personas on LLMs' performance vary with the role categories? To answer this question, we run a mixed effect model using the personas' role category to predict the prediction accuracy for each question, controlling the model as a random effect. We found that work- and school-related roles are associated with better performances than other types of roles, especially AI and occupational roles. However, the effect sizes are relatively small, suggesting that 
the category of persona does not have a large effect on LLMs' performances. 

\begin{figure}[t]
    \centering    
    \includegraphics[width=\columnwidth]{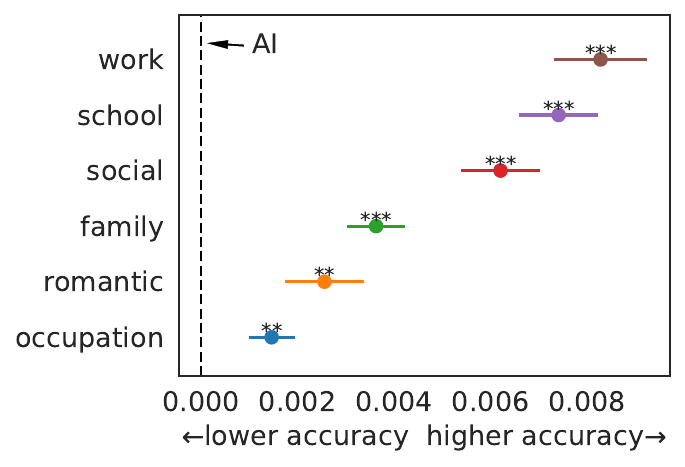}
    \caption{Work- and School-related Roles lead to better performances than other types of roles across models.}
    \label{fig:lme-role-cate}
\end{figure}

\paragraph{Domain Alignment}

\begin{figure*}[!ht]
    \centering
    \begin{subfigure}{0.32\textwidth}
        \includegraphics[width=\linewidth]{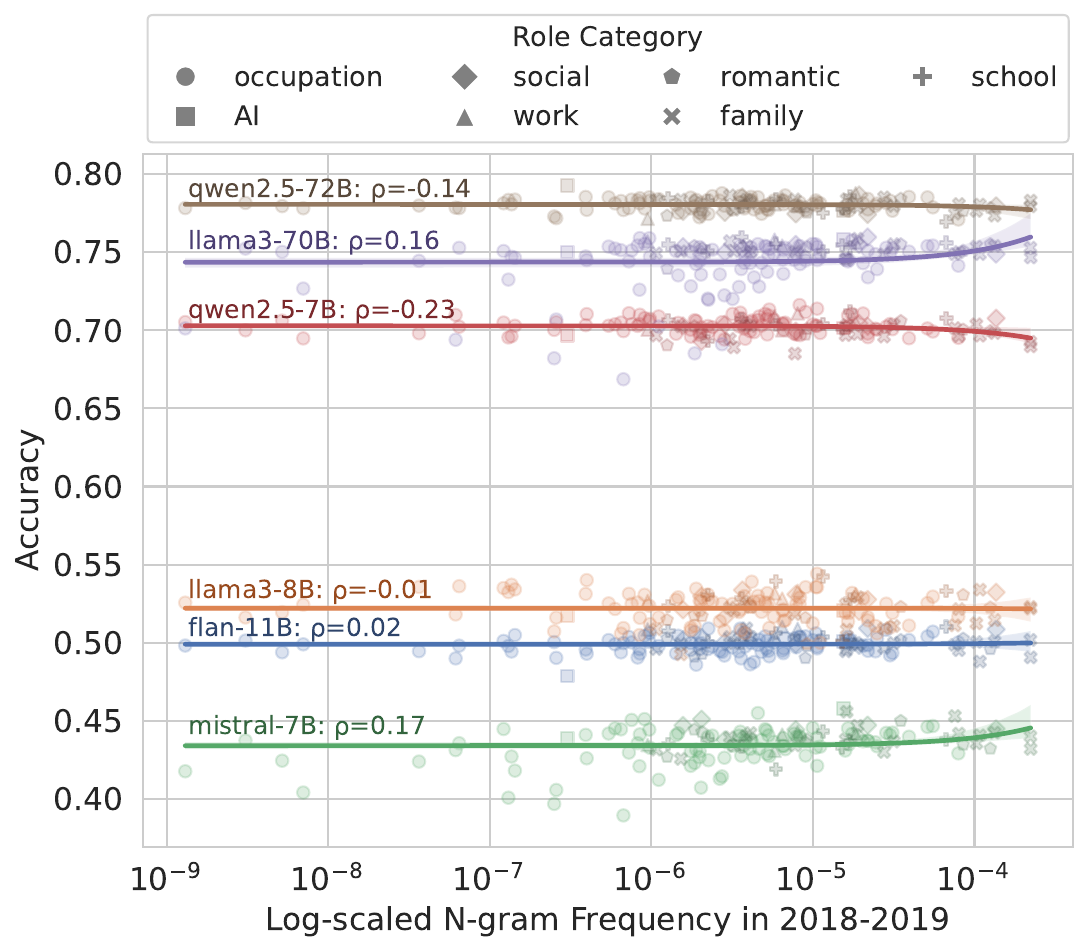}
        \caption{Word frequency}
        \label{fig:freq_overall_cate_shaped}
    \end{subfigure}
    \begin{subfigure}{0.32\textwidth}
        \includegraphics[width=\linewidth]{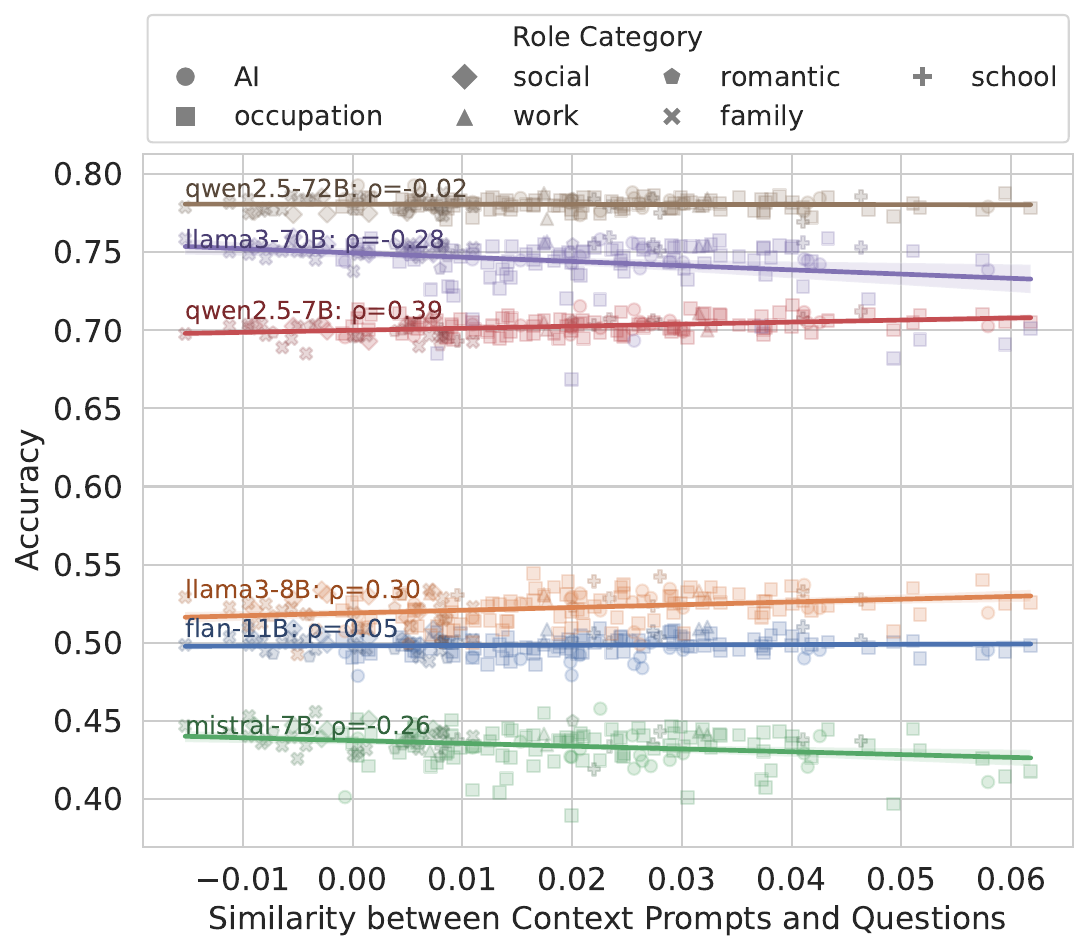}
        \caption{Prompt-question similarity}
        \label{fig:sim_overall_cate_shaped}
    \end{subfigure}
    \begin{subfigure}{0.32\textwidth}
        \includegraphics[width=\linewidth]{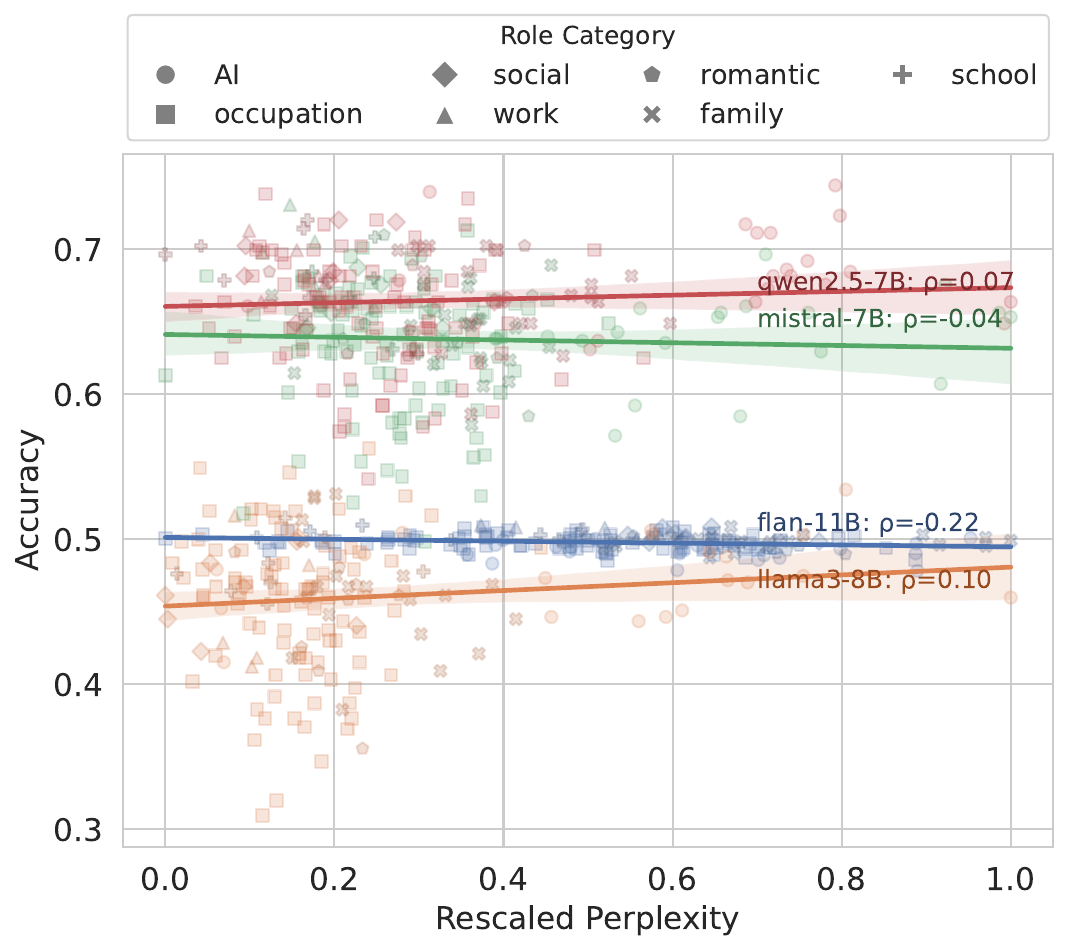}
        \caption{Prompt perplexity}
        \label{fig:ppl_role_cate_shaped}
    \end{subfigure}
    \caption{(a) personas' word frequency is weakly correlated with model performance. (b) prompt-question similarity shows weak to moderate correlations with the models' performance. (c) The perplexity of the prompt has a negative and weak correlation with the models' performance.}
    \label{fig:explain-role-diff-all}
\end{figure*}

While we observe no significant differences between most of the personas and the control setting, it is possible that certain roles might still lead to better answers for specific questions. For example, many prompt engineering guidebooks suggest adding roles that are aligned with the current conversation context \footnote{\url{https://llama.meta.com/docs/how-to-guides/prompting/}}. Do domain-aligned personas really lead to better model performance? To test this question, we label each role-question pair with ``in-domain'' and ``out-domain'' based on its category. For example, if the persona is ``software engineer'' and the question is in Computer Science, we consider it as an in-domain pair.

To assess the effect of domain alignment, we fit another mixed-effects model using the binary in-domain indicator as the sole predictor and include a random effect for each model. The coefficient for ``in-domain`` is 0.004 (p < 0.01), suggesting that in-domain roles generally lead to better performances than out-domain roles. For example, lawyers are more likely to give accurate answers to law-related questions than doctors. However, the effect size of domain alignment is relatively small, suggesting that the domain alignment between the personas and questions only has minor effects on LLMs' performance on objective tasks.

\begin{figure*}[t]
    \centering
    \begin{subfigure}{0.45\textwidth}
        \centering
        \includegraphics[width=\linewidth]{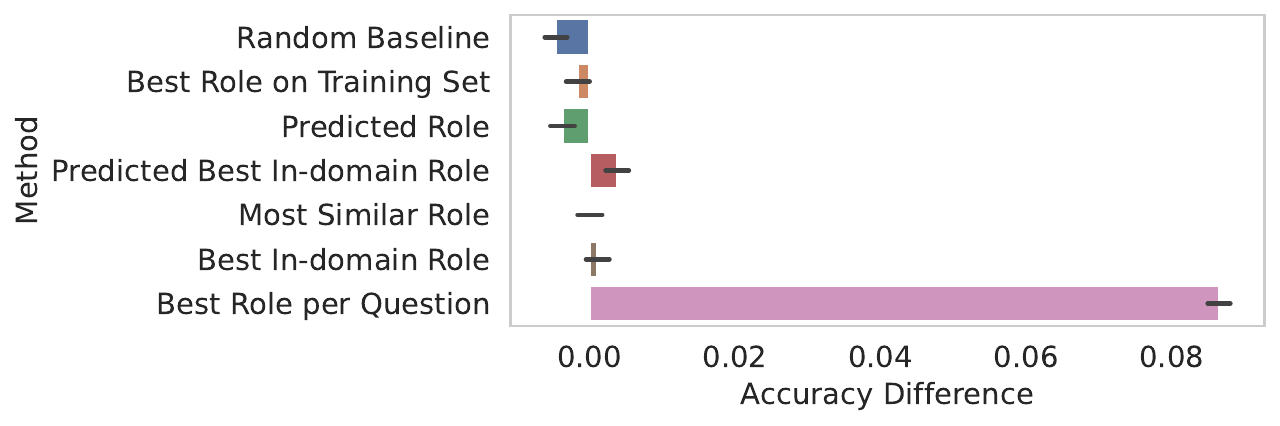}
        \caption{FLAN-T5-11B}
        \label{fig:flan-all}
    \end{subfigure}
    \begin{subfigure}{0.45\textwidth}
        \centering
        \includegraphics[width=\linewidth]{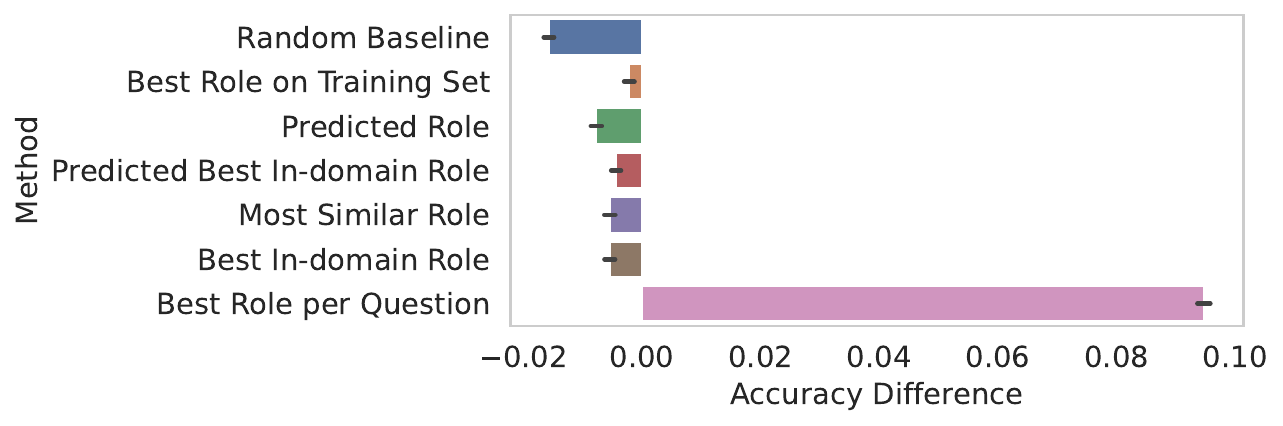}
        \caption{Llama3-70B}
        \label{fig:llama-all}
    \end{subfigure}
    \\
    \begin{subfigure}{0.45\textwidth}
        \centering
        \includegraphics[width=\linewidth]{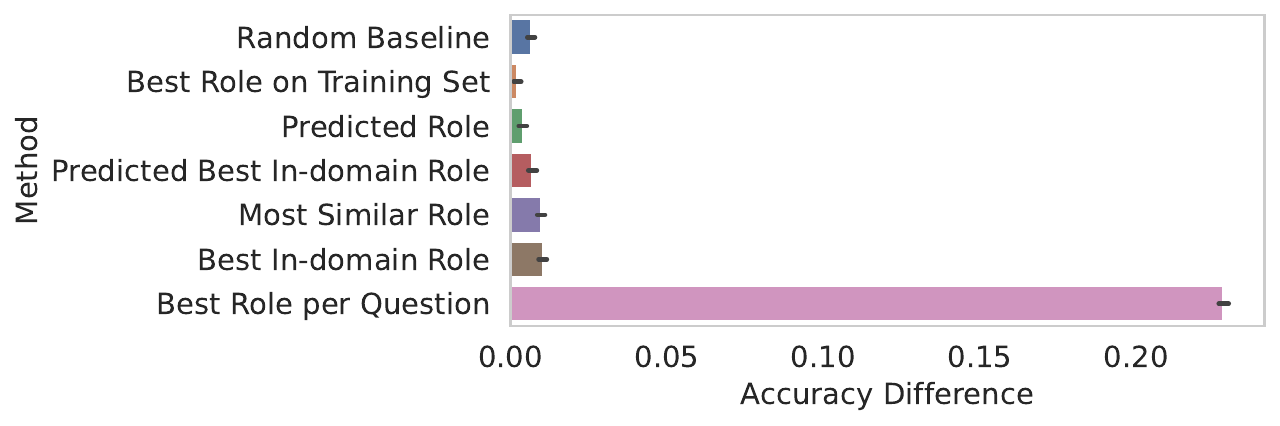}
        \caption{Mistral-7B}
        \label{fig:mistral-all}
    \end{subfigure}
    \begin{subfigure}{0.45\textwidth}
        \centering
        \includegraphics[width=\linewidth]{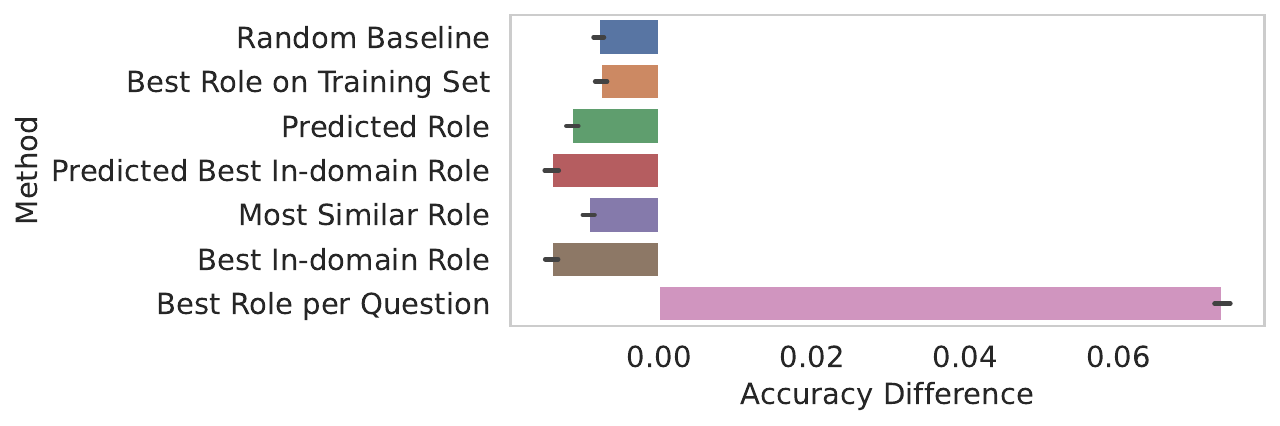}
        \caption{Qwen2.5-72B}
        \label{fig:qwen-all}
    \end{subfigure}
    \caption{Performance change for each model (compared with the control prompt) across different role-selection strategies reveals that the best-performing role per question is often idiosyncratic and different strategies for selecting the appropriate role offer limited (if any) improvement over picking a random role.}
    \label{fig:role-pick-all-models}
\end{figure*}

\section{Why Certain Personas Lead to Higher Accuracies?}
 
Why do certain personas lead to better performance than others? Despite the complexity across personas, we assess several potential mechanisms. In this section, we propose a method to calculate persona embedding that enables an overall performance comparison. Furthermore, we test whether specific characteristics of the prompt and personas might be driving the behavior: the n-gram frequency of role words, the similarity between context prompts and questions, and the perplexity of the context prompts.

\paragraph{Word Frequency of Personas}

For each role, we obtain its n-gram frequency for the period between 2018 and 2019 (the most recent data available) from the Google Ngram Viewer \footnote{\url{https://books.google.com/ngrams/}}. Frequencies are retrieved for the full role n-gram (e.g., the frequency of ``software engineer'' as a bigram). Figure~\ref{fig:freq_overall_cate_shaped} illustrates the aggregated relationship between accuracy and role word frequency for each model, where each point represents a role and is characterized by its role category. The n-gram frequencies of roles are weakly correlated with their accuracy across all models, as indicated by the Pearson correlation coefficients. The largest correlation in absolute value is -0.23 for Qwen2.5-7B, and the smallest is -0.01 for Llama3-8B. This trend suggests that word frequency does not fully explain the effect of personas on model performance. 

\paragraph{Prompt-Question Similarity} All of our prompts include two parts: context (e.g., You are talking to your boss) and questions (e.g., Where is the capital of the United States?). Are context prompts that are semantically more similar to the questions more likely to generate accurate answers? To answer this question, we utilize MiniLM \citep{wang2020minilm} from Sentence-BERT package \citep{reimers2019sentence} to encode a set of context prompts and full questions with options, and then compute the cosine similarity between the two vectors as a measure of distance between the question and prompt. As shown in Figure~\ref{fig:sim_overall_cate_shaped}, we observe a weak correlation between similarity and accuracy at the role level. Furthermore, the effect of similarity is inconsistent across different model families and sizes. Specifically, the highest correlation is 0.39 on Qwen2.5-7B, whereas the smallest absolute correlation is -0.02 for Qwen2.5-72B. 

\paragraph{Prompt Perplexity}

Perplexity quantifies the overall probability of a piece of text for a given language model. It serves as an indicator of the model's uncertainty, with lower perplexity indicating more common sequences. We use each model's tokenizer and architecture to compute model-specific perplexities. For FLAN-T5, we use a pair of context prompts and the questions as the input. For other models, perplexity is computed for the entire prompt, consisting of a context prompt followed by a question with options.  We further rescale the calculated perplexity scores to a range of 0 to 1 to allow easier comparisons across models. As shown in Figure~\ref{fig:ppl_role_cate_shaped}, the mean accuracy is negatively correlated with the rescaled perplexity at the role level on FLAN-T5 and Mistral, whereas the correlation is positive on Llama and Qwen. These results suggest that logical coherence and inherent reasonability of prompts do not necessarily result in more accurate responses. The impact of perplexity is model-dependent as well. 

\paragraph{Overall Regression Analysis} To perform a comprehensive analysis of all the attributes of roles mentioned previously, we fit a mixed-effects model using three independent variables: the role's n-gram frequency, prompt-question similarity, and prompt-question perplexity. Random intercepts are included for each model. \tref{tab:lme_overall} in the Appendix details the regression results. We find that higher frequency, higher similarity, and lower perplexity are generally associated with higher prediction accuracy ($p < 0.05$), reflecting similar patterns as shown in the correlation analysis.

\section{Finding the Best Personas for Prompting}

In previous sections, we demonstrate that there might not exist a single persona that consistently improves the performance of diverse sets of questions. However, we also observe that personas might help in cases where their domains are aligned with the questions or when they have higher similarities.  A natural question arises: instead of using the same role for all questions, could we automatically find the best role for prompting in a specific setting? We experiment with a set of methods to select the best role for prompting LLMs. For the four LLM collections used in our analysis, we choose the model with the largest number of parameters for automatic role search experiments. 

\subsection{Methods}
We experiment with the following baselines in selecting the best roles for prompting.
\textbf{Random:} Randomly select a role from the predefined role list for each question. 
\textbf{In-domain best role:} Automatically select the best in-domain role in the training set. 
\textbf{Best role:} Automatically select the best role in the training data. 
\textbf{Best role per question:} Automatically select the best role per question in the test data, this is the performance upper bound.

We further design the following methods to automatically select the best roles. \textbf{Similarity-based method:} Select the role that has the highest similarity to the question. \textbf{Dataset classifier: } A classifier aims at finding the correct domain for each question. We first fine-tune a \texttt{roberta-base} model to predict the domain of the question. We concatenate the entire question with its options as the input, and the output is the domain of the question. We further select the best in-domain role from the training set. The 2,410 questions are divided into a 7:1:2 ratio for training, validation, and the test set, respectively. The overall accuracy of the domain classifier is 78.1\% on the test set. For reference, the accuracies of a random guess and choosing the most frequent class are 5.2\% and 6.9\%, respectively. \textbf{Role Classifier:} A classifier aims at predicting the best role for each question. To this end, we fine-tune a \texttt{roberta-base} model and use it as a multi-label classifier for personas. The prediction target is the 162 roles, and the classifier achieved an F1 score of 0.34 for FLAN-T5-XXL, 0.39 for Mistral-7B, 0.71 for Llama3-70B, and 0.77 for Qwen2.5-72B on the test set.

\subsection{Results}

Figure~\ref{fig:role-pick-all-models} shows overall model performance using different role-searching strategies on four models relative to the control group (i.e., prompting with no role). The best role per question can be considered as the upper bound for the persona effect on model performance, where the model accurately picks the best role for each question. We find that when automatically selecting the best role, the aggregated result can generally lead to significantly better overall performance. This suggests that for each specific question, there exist certain personas that can lead to the correct answer. 
However, all of the automatic role-searching strategies are far away from this upper bound. Furthermore, most role-searching strategies are just marginally better than randomly selecting a role for each question---and for the Qwen model, these strategies do worse than random.  This result suggests that the effect of personas on LLMs' performances might naturally be unpredictable.

\section{Conclusion}
Incorporating personas into prompts has been an important approach for designing system prompts and enabling role-playing with LLMs. However, the effects of adding these personas on model performance were previously unclear. In this study, we present a systematic analysis of 162 personas in 26 categories to explore how prompting with personas affects model performance. Our analysis shows that, compared with the control setting, adding a persona does not necessarily improve an LLM’s performance on objective tasks. On the contrary, it might actually hurt the models' overall performance in some situations. Furthermore, while a specific persona may lead to the correct answer for individual questions, and aggregating these personas can result in significant performance gains, identifying the best role remains challenging, as most selection strategies perform similarly to random selection. This result suggests that the effect of personas on model performance can be largely unpredictable. Our study introduces a new computational pipeline to evaluate the impact of adding personas on LLM performance. These findings can help inform the future design of system prompts and role-playing strategies with LLMs. All data, results, and experiment code are available at \url{https://github.com/Jiaxin-Pei/Prompting-with-Social-Roles}.

\section*{Acknowledgments}
This work was funded in part by LG AI Research and by the National Science Foundation under Grant No. IIS-2143529. The authors thank Honglak Lee for his feedback on early versions of this work.

\section{Limitations}
Our study has the following limitations: First, we only studied four open-source LLM families and didn't include closed-source models like GPT3.5 and GPT4. This is due to the computational cost of running such a large experiment. We will release the script to run the experiment and we welcome other researchers to explore how role-playing affects LLM performance on other models.  Second, while we aimed to be comprehensive in selecting personas, we were unable to experiment with all possible personas beyond the 162 used in our current experiment. We will release the full list of personas to support future research in this area. Third, due to the computational costs of our experiments, we used only MMLU as our testbed, overlooking other factual question datasets and open-ended questions. While we believe our current analysis offers important insights into how personas affect model performance, we acknowledge this limitation and plan to extend our analysis to additional settings. 

\section{Ethical Considerations}
Our study has the following ethical implications. First, to ensure the robustness of our results, we experimented with 162 roles, 4 prompt templates, and 9 LLMs across 2,410 MMLU questions. Running such experiments is computationally expensive and likely contributes to a substantial carbon dioxide footprint. Second, some of our analyses may reinforce existing stereotypes regarding personas. For example, our results suggest that masculine roles lead to better performance than feminine roles, which might inadvertently reinforce traditional gender stereotypes. However, our results indicate that gender-neutral roles result in higher performance than gendered roles, suggesting that developers should prioritize gender-neutral roles when creating system prompts. Additionally, our results reveal potential model biases stemming from implicit societal stereotypes regarding gender roles. We call for future research to study de-biasing technologies when training or aligning LLMs.

\bibliographystyle{acl_natbib}
\bibliography{custom}

\appendix
\section{Experiment Settings}

\paragraph{Dataset and Models} 

The dataset and models used in this study, along with their licenses, are listed in \tref{tab:model_license}. All of them are open-source, and our use aligns with their intended purpose. The mapping between sampled MMLU subsets and their domains is shown in \tref{tab:domain_mapping}. 

\begin{table}[ht]
    \centering
    \resizebox{\linewidth}{!}{   
    \begin{tabularx}{\columnwidth}{lX}
        \toprule
        \textbf{Model/Dataset} & \textbf{License} \\
        \midrule
        MMLU & MIT \\
        \midrule
        Flan-T5 & Apache-2.0 \\
        \midrule
        Llama-3 & llama3 \\
        \midrule
        Mistral-v0.2 & Apache-2.0 \\
        \midrule 
        Qwen2.5 & Apache-2.0\\
        \bottomrule
    \end{tabularx}}
    \caption{List of licenses}
    \label{tab:model_license}
\end{table}

\begin{table*}[h!]
    \centering
    \begin{tabularx}{\textwidth}{lX}
        \toprule
        Domain & Datasets \\
        \midrule
        Law & professional\_law, international\_law \\
        Medicine & clinical\_knowledge, college\_medicine, professional\_medicine \\
        EECS & electrical\_engineering, college\_computer\_science, high\_school\_computer\_science \\
        Math & high\_school\_statistics, college\_mathematics, high\_school\_mathematics \\
        Politics & us\_foreign\_policy, high\_school\_government\_and\_politics \\
        Psychology & professional\_psychology, high\_school\_psychology \\
        Natural Science & college\_physics, college\_biology, high\_school\_physics, high\_school\_chemistry, college\_chemistry, high\_school\_biology \\
        Econ & management, professional\_accounting, econometrics, high\_school\_macroeconomics, high\_school\_microeconomics \\
        \bottomrule
    \end{tabularx}
    \caption{Domain Dictionary}
    \label{tab:domain_mapping}
\end{table*}

\paragraph{Roles and Prompts} The full list of roles is shown in \tref{tab:all_roles}, and the roles used for explicit gender impact are listed in \tref{tab:aligned_gender_roles}. The 4 prompt templates are listed in \tref{tab:prompts} and the deailed context prompts and control prompts are shown in \tref{tab:context_prompts} and \tref{tab:control_prompts}. 

\begin{table*}[ht]
    \centering
    \begin{tabularx}{\textwidth}{lX}
        \toprule
        Category & Roles \\
        \midrule
        family & sister, son, father-in-law, mother-in-law, brother, parent, father, mother, daddy, dad, papa, mummy, mamma, mommy, mom, mum, mama, daughter, cousin, grandfather, grandmother \\
        romantic & partner, husband, wife, boyfriend, housewife, girlfriend, fiancée, fiancé \\
        school & professor, instructor, student, coach, tutor, dean, graduate, classmate\\
        work & supervisor, coworker, boss, colleague, mentor \\
        social & companion, buddy, roommate, friend, stranger, foreigner, best friend, close friend \\
        AI & chatbot, assistant, virtual assistant, AI language model, mathematician AI, software engineer AI, Educational Tutor AI, Medical Diagnostic AI, helpful assistant, Behavioral Economics AI, Historical Data Analyst AI, Legal Research AI, Mathematical Modeling AI, Statistical Analysis AI, Diagnostic AI, Policy Analysis AI, Public Opinion AI, Psychological Profiling AI, Scientific Data Analysis AI, Embedded Systems AI Engineer\\
        econ & economic researcher, economist, financial analyst \\
        eecs & electronics technician, data scientist, electrical engineer, software engineer, web developer\\
        history & historian, archivist, historical researcher, archaeologist\\
        law & bailiff, lawyer\\
        math & data analyst, mathematician, statistician\\
        medicine & nurse, doctor, physician, dentist, surgeon\\
        natural science & geneticist, biologist, physicist, teacher, chemist, ecologist\\
        other occupations & painter, auctioneer, musician, scientist, driver, accountant, geologist, janitor, architect, mason, baker, administrator, research scientist, weaver, postmaster, cook, clerk, broker, dancer, surveyor, clergy, secretary, soldier, housekeeper, collector, carpenter, cashier, conductor, mechanic, engineer, photographer, manager, farmer, tailor, shoemaker, sales, librarian, blacksmith, artist, pilot, inspector, police, gardener, attendant, athlete, operator, sailor, designer, midwife, president, humanist, auditor, scholar, CEO, advisor, counsellor, counselor, cofounder\\
        politics & politician, sheriff, governer, enthusiast, partisan\\
        psychology & psychologist\\
        \bottomrule
    \end{tabularx}
    \caption{Role Dictionary}
    \label{tab:all_roles}
\end{table*}

\begin{table}[ht]
    \centering
    \resizebox{\linewidth}{!}{%
    \begin{tabularx}{\columnwidth}{lX}
        \toprule
        \textbf{Gender} & \textbf{Roles} \\
        \midrule
        Masculine & father, daddy, dad, papa, father-in-law, grandfather, husband, son, boyfriend, fiancé \\
        \midrule
        Feminine & mother, mommy, mom, mamma, mother-in-law, grandmother, wife, daughter, girlfriend, fiancée \\
        \midrule
        Neutral & partner, parent\\
        \bottomrule
    \end{tabularx}
    }
    \caption{List of aligned roles categorized by gender}
    \label{tab:aligned_gender_roles}
\end{table}

\begin{table}[ht]
    \centering
    \begin{tabularx}{\columnwidth}{lX}
        \toprule
        \textbf{Prompt Type} & \textbf{Prompt} \\
        \midrule
        Audience-Specific & 
        You are talking to a/an \{role\}. \\
        & Imagine you are talking to a/an \{role\}.\\
        \midrule
        Speaker-Specific & 
        You are a/an \{role\}.  \\
        & Imagine you are a/an \{role\}. \\
        \bottomrule
    \end{tabularx}
    \caption{Context prompts}
    \label{tab:prompts}
\end{table}

\begin{table*}[ht]
    \centering
    \resizebox{\textwidth}{!}{%
    \begin{tabular}{|p{2cm}|p{14cm}|}
        \hline
        \textbf{Model Type} & \textbf{Prompt Template} \\
        \hline
        FLAN-T5 & 
        \{context\_prompt\} \{question\} \ Please select the correct answer number: \\
        \hline
        Llama3, \newline
        Mistral, \newline
        Qwen2.5 & 
        \{``role'': ``system'', ``content'': \{context\_prompt\}\}, \newline 
        \{``role'': ``user'', ``content'': The following is a multiple choice question (with answers). Reply with only the option number. \{question\}\} \\
        \hline
    \end{tabular}}
    \caption{Context Prompts for each model}
    \label{tab:context_prompts}
\end{table*}

\begin{table*}[ht]
    \centering
    \resizebox{\textwidth}{!}{%
    \begin{tabular}{|p{2cm}|p{14cm}|}
        \hline
        \textbf{Model Type} & \textbf{Prompt Template} \\
        \hline
        FLAN-T5 & 
        \{question\} Please select the correct answer number: \\
        \hline
        Llama3, \newline
        Mistral, \newline
        Qwen2.5 &  
        \{``role'': ``user'', ``content'': The following is a multiple choice question (with answers). Reply with only the option number. \{question\}\} \\
        \hline
    \end{tabular}}
    \caption{Control Prompts for each model}
    \label{tab:control_prompts}
\end{table*}

\section{Regression Results}
\label{appendix:role-regression}

\begin{table}[ht]
    \centering
    \resizebox{\linewidth}{!}{%
    \begin{tabularx}{\columnwidth}{lXX}
        \toprule
        \textbf{Term} & \textbf{Coefficient} & \textbf{p-value} \\
        \midrule
        Frequency & 108.638 & 3.12e-02 \\
        \midrule
        Perplexity & -0.000398 & 2.45e-02 \\
        \midrule
        Similarity & 0.419 & 3.66e-64 \\
        \bottomrule
    \end{tabularx}
    }
    \caption{Coefficients of the mixed-effects model on the relationship between accuracy and all the role attributes}
    \label{tab:lme_overall}
\end{table}

\begin{figure*}[ht]
    \centering    \includegraphics[width=0.50\textwidth]{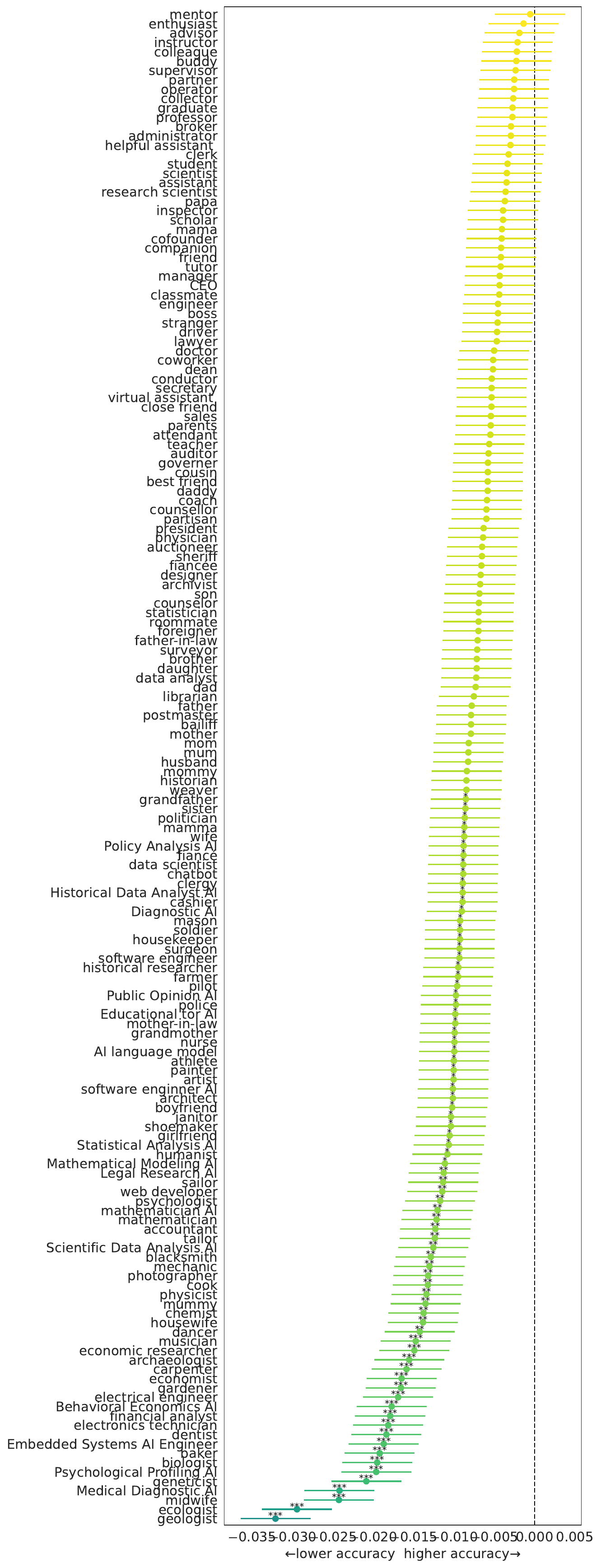}
    \caption{Coefficients of the regression model on the relationship between accuracy and role with random effects for each model}
    \label{fig:acc-role-mlm-full}
\end{figure*}

\begin{figure*}[ht]
    \centering    \includegraphics[width=0.50\textwidth]{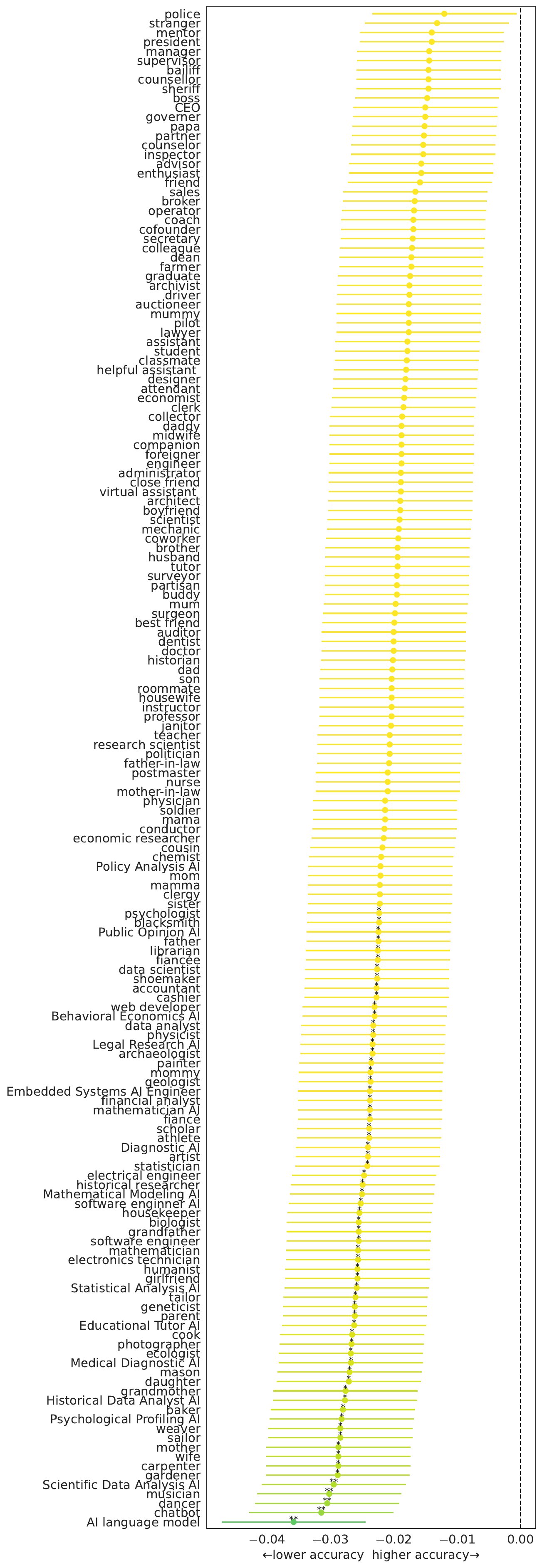}
    \caption{Coefficients of the regression model on the relationship between accuracy and role with random intercepts for Flan-T5-XXL}
    \label{fig:acc-role-flan-full}
\end{figure*}

\begin{figure*}[ht]
    \centering    \includegraphics[width=0.50\textwidth]{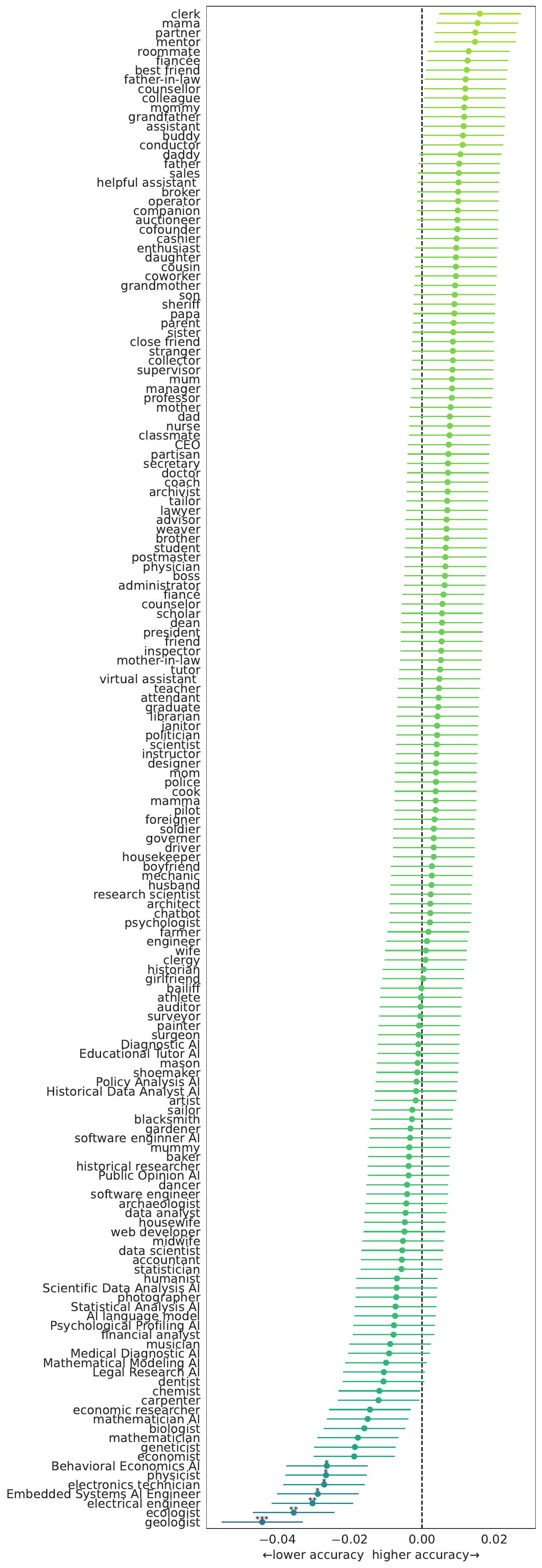}
    \caption{Coefficients of the regression model on the relationship between accuracy and role with random intercepts for Mistral-7B}
    \label{fig:acc-role-mistral-full}
\end{figure*}

\begin{figure*}[ht]
    \centering    \includegraphics[width=0.50\textwidth]{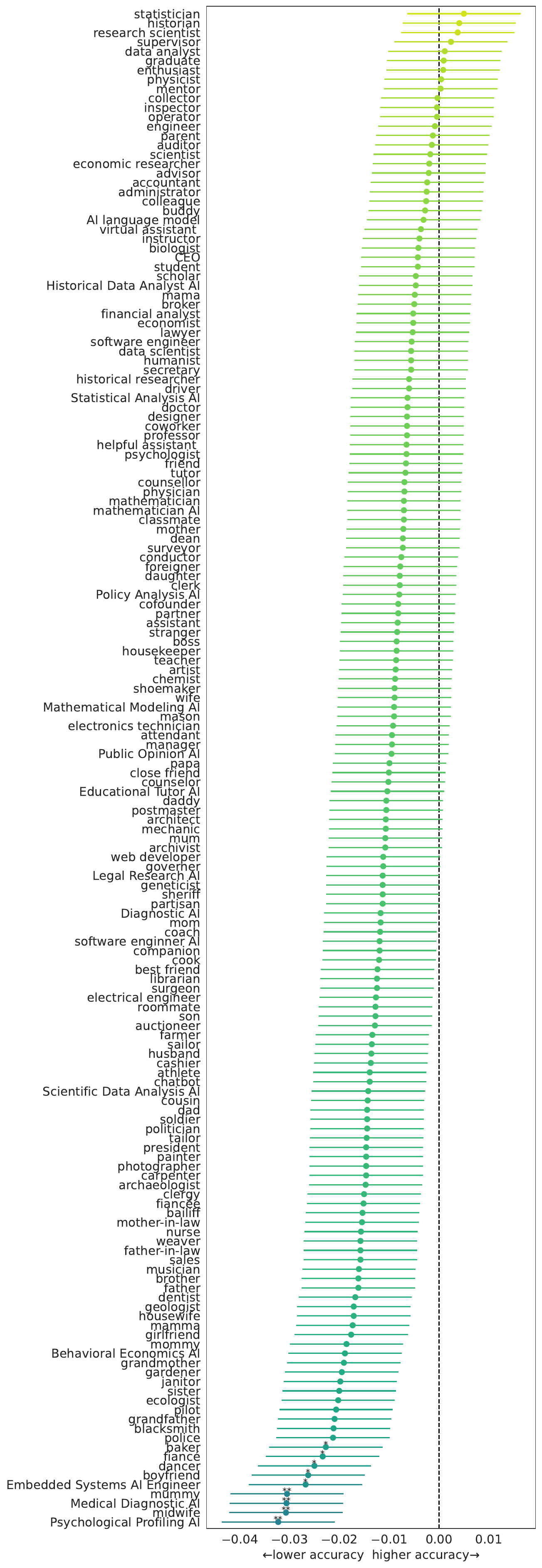}
    \caption{Coefficients of the regression model on the relationship between accuracy and role with random intercepts for Llama-3-8B}
    \label{fig:acc-role-llama-8B}
\end{figure*}

\begin{figure*}[ht]
    \centering    \includegraphics[width=0.50\textwidth]{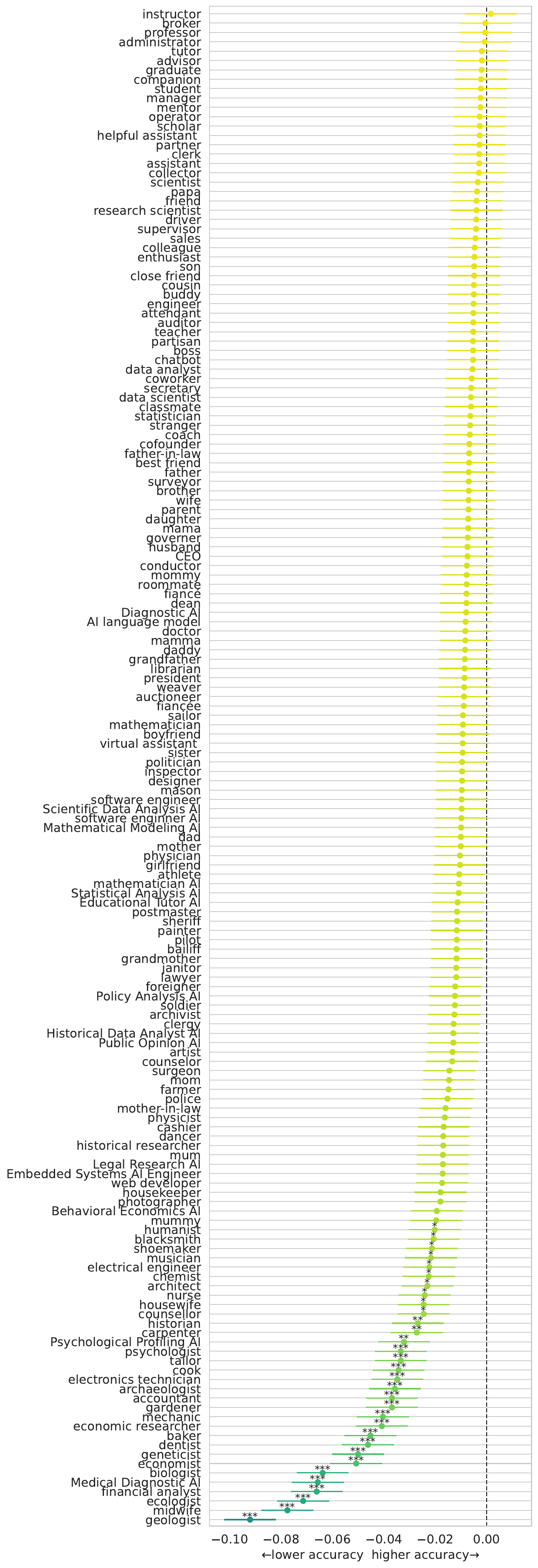}
    \caption{Coefficients of the regression model on the relationship between accuracy and role with random intercepts for Llama-3-70B}
    \label{fig:acc-role-llama-70B}
\end{figure*}

\begin{figure*}[ht]
    \centering    \includegraphics[width=0.50\textwidth]{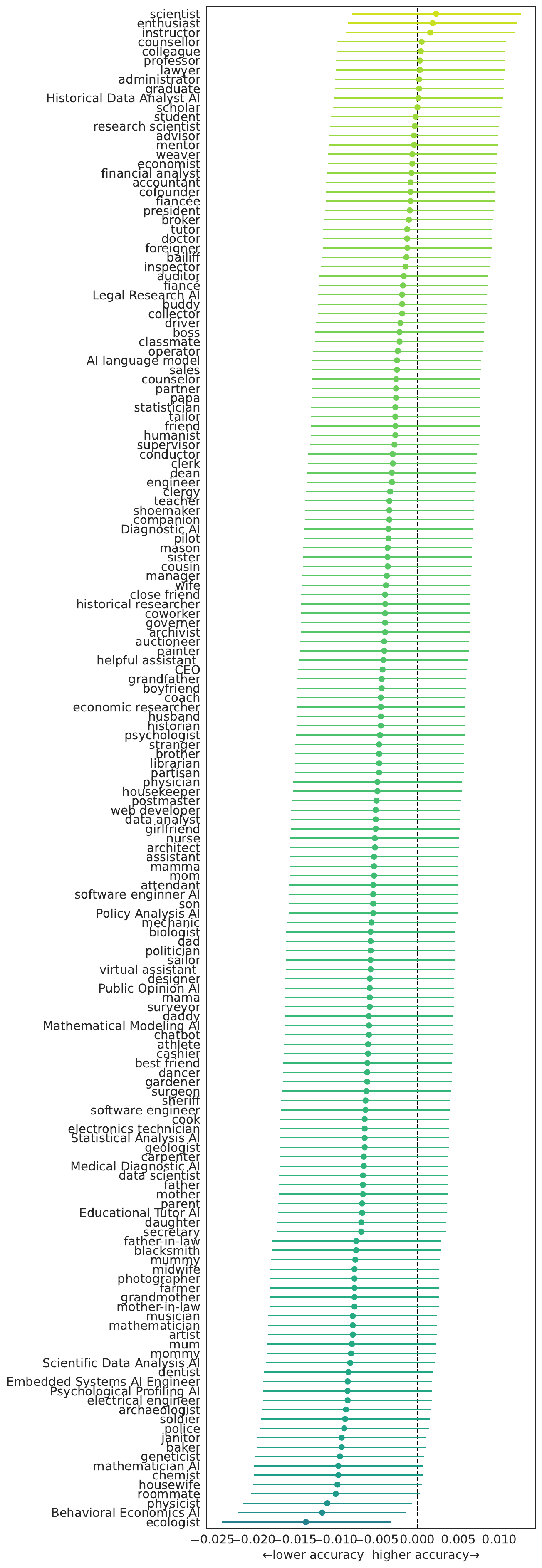}
    \caption{Coefficients of the regression model on the relationship between accuracy and role with random intercepts for Qwen2.5-7B}
    \label{fig:acc-role-qwen-7B}
\end{figure*}

\begin{figure*}[ht]
    \centering    \includegraphics[width=0.50\textwidth]{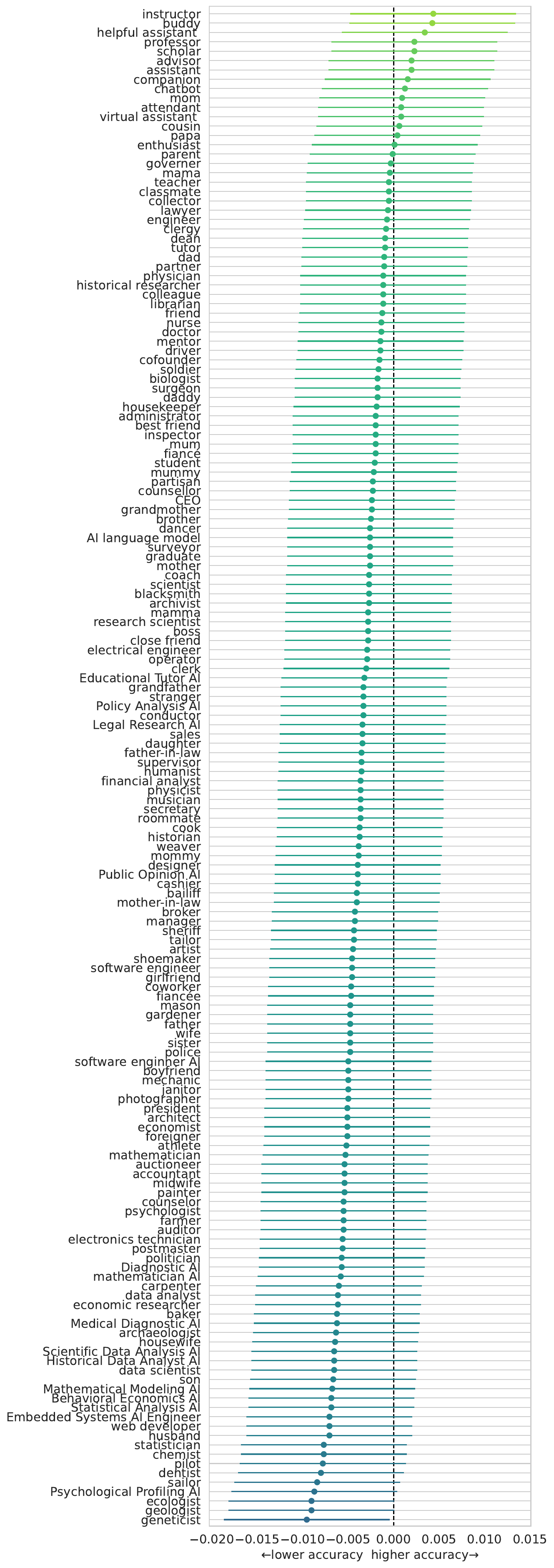}
    \caption{Coefficients of the regression model on the relationship between accuracy and role with random intercepts for Qwen2.5-72B}
    \label{fig:acc-role-qwen-72B}
\end{figure*}

\paragraph{Persona Impact} Figure~\ref{fig:acc-role-mlm-full} shows the regression coeffects of ``role'' when controlling a random effect for each model. Figures~\ref{fig:acc-role-flan-full}, \ref{fig:acc-role-mistral-full}, \ref{fig:acc-role-llama-8B}, \ref{fig:acc-role-llama-70B}, \ref{fig:acc-role-qwen-7B} and \ref{fig:acc-role-qwen-72B} show the coefficients of ``role'' in the linear relationship between accuracy and role for each model.   

\paragraph{Overall Regression} \tref{tab:lme_overall} lists the coefficients and p-values for the mixed-effects model on the impact of frequency, similarity and perplexity on prediction accuracy, controlling for each model as a random effect. 

\section{Persona Embeddings}

\begin{figure*}[ht]
    \centering    \includegraphics[width=\textwidth]{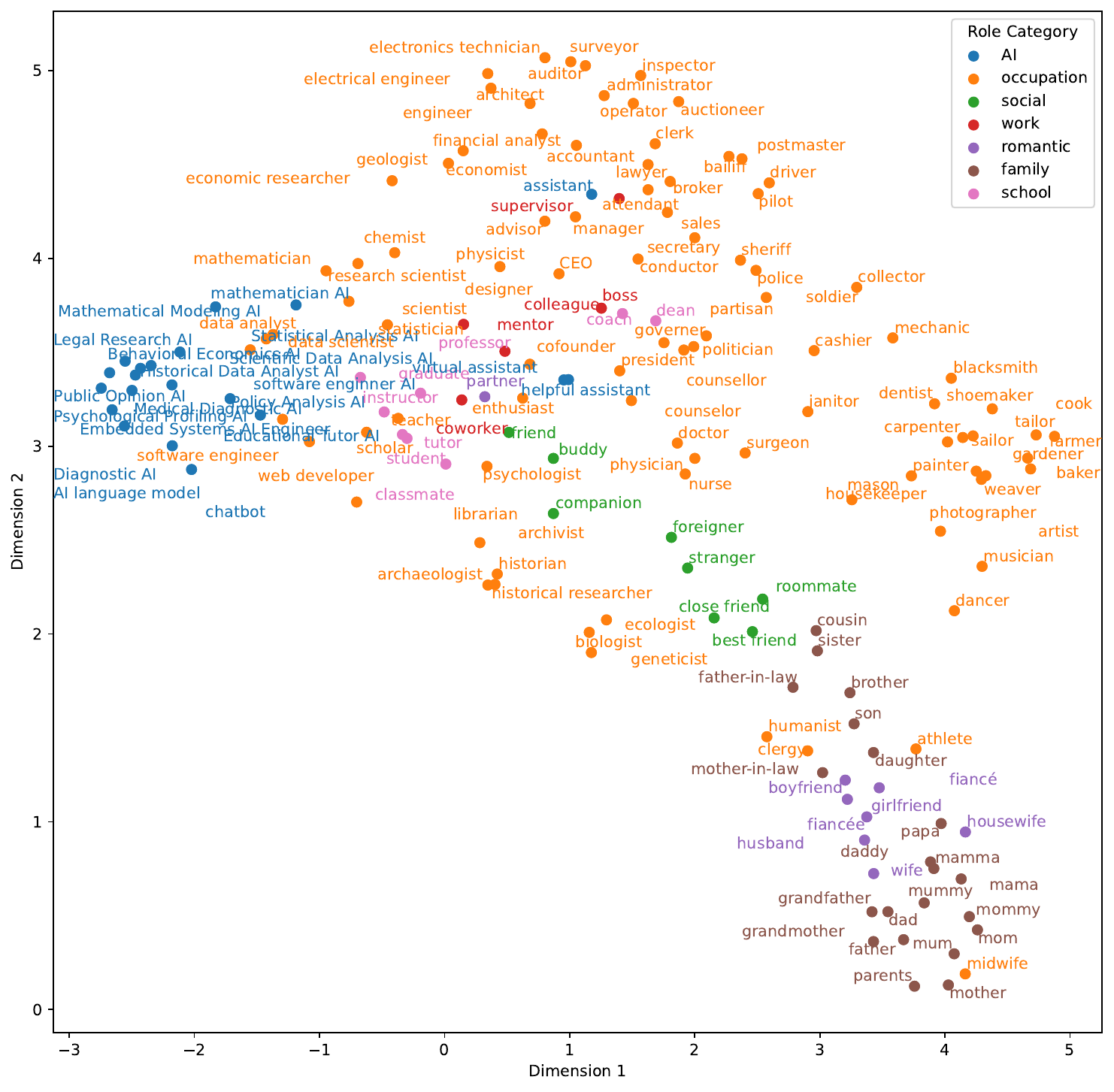}
    \caption{Role embeddings calculated by UMAP for Flan-T5-XXL}
    \label{fig:role-embed-flan}
\end{figure*}

\begin{figure*}[ht]
    \centering    \includegraphics[width=\textwidth]{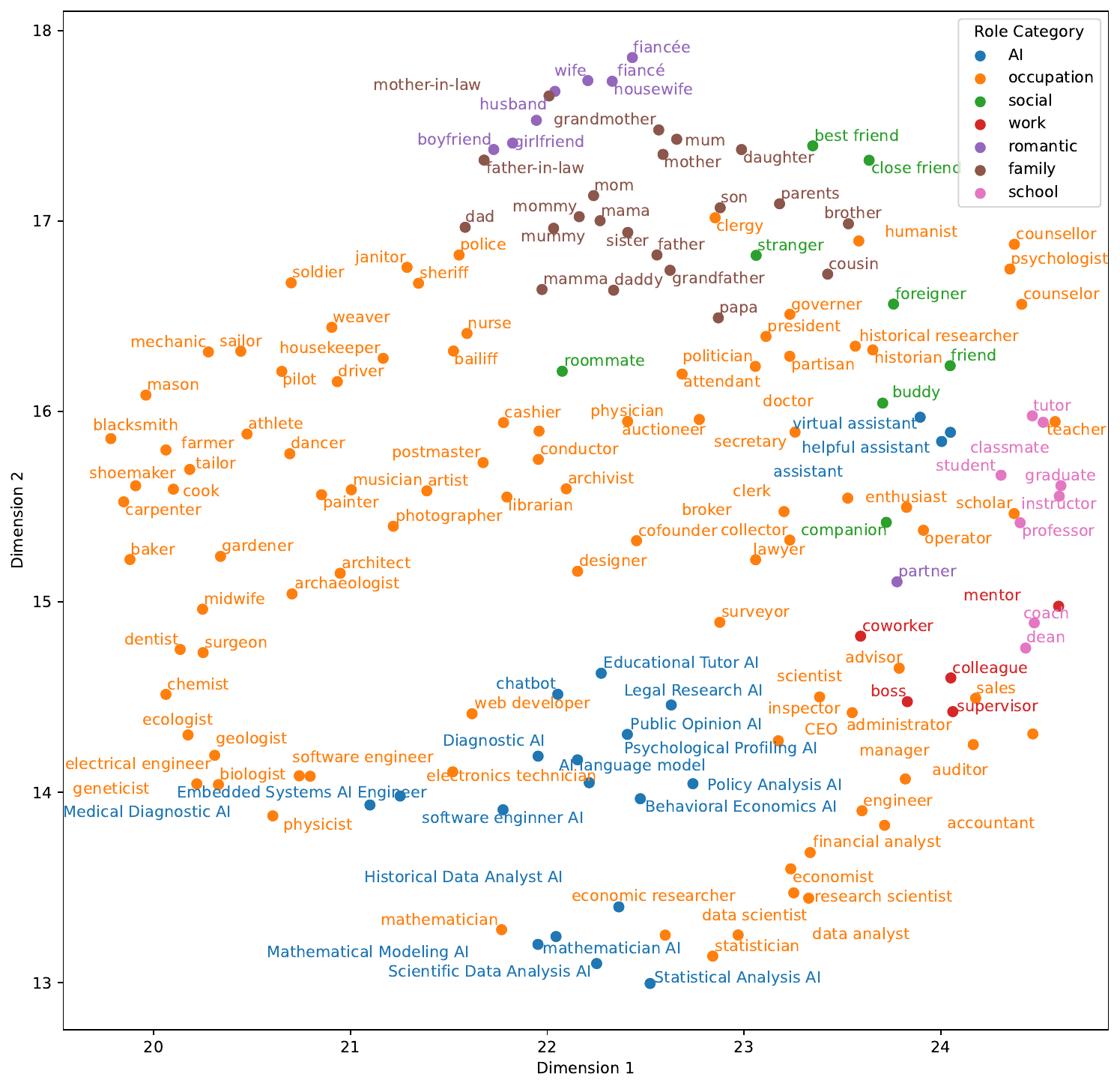}
    \caption{Role embeddings calculated by UMAP for Mistral-7B}
    \label{fig:role-embed-mistral}
\end{figure*}

\begin{figure*}[ht]
    \centering    \includegraphics[width=\textwidth]{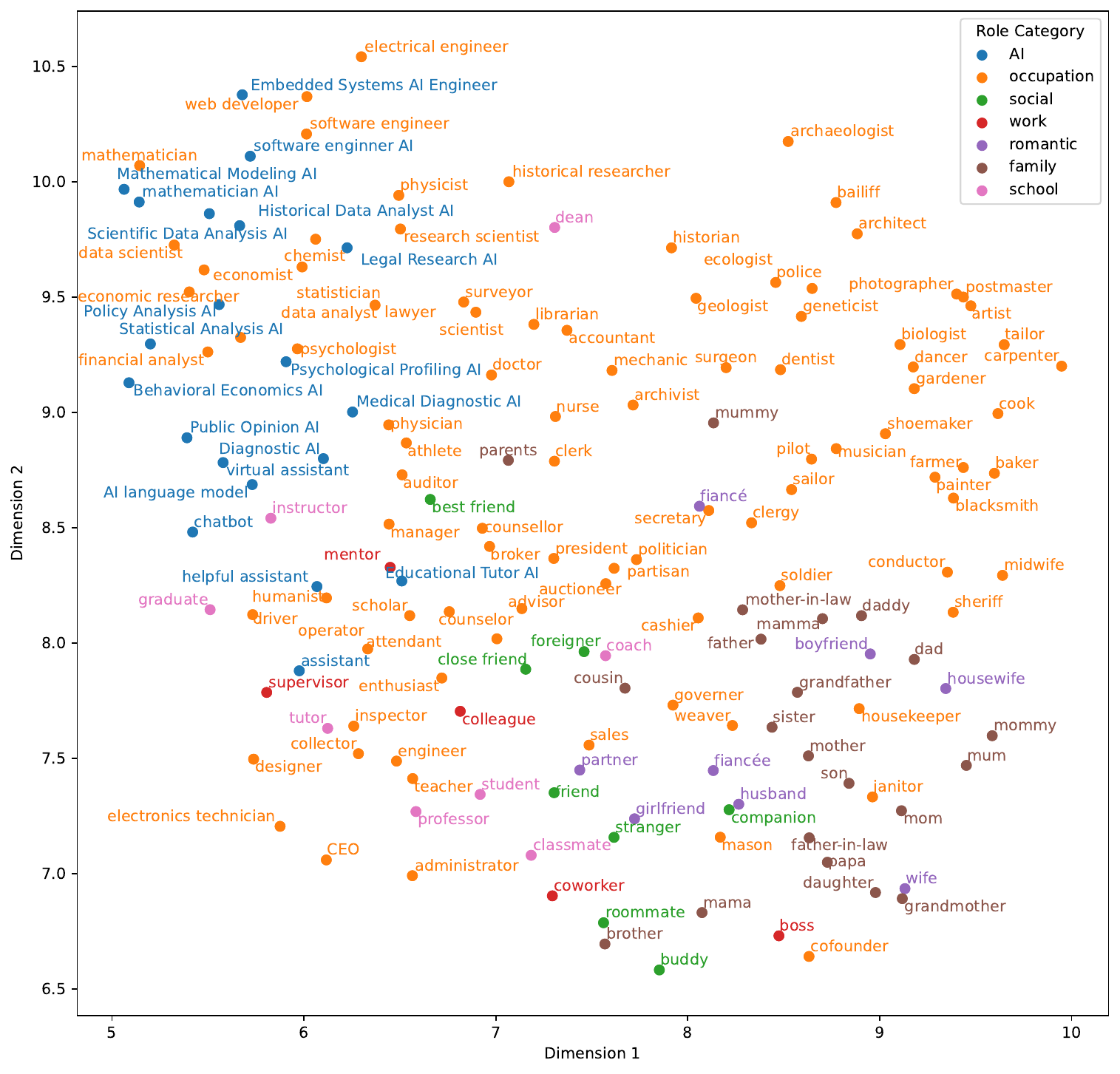}
    \caption{Role embeddings calculated by UMAP for Llama3-8B}
    \label{fig:role-embed-llama-8B}
\end{figure*}

\begin{figure*}[ht]
    \centering    \includegraphics[width=\textwidth]{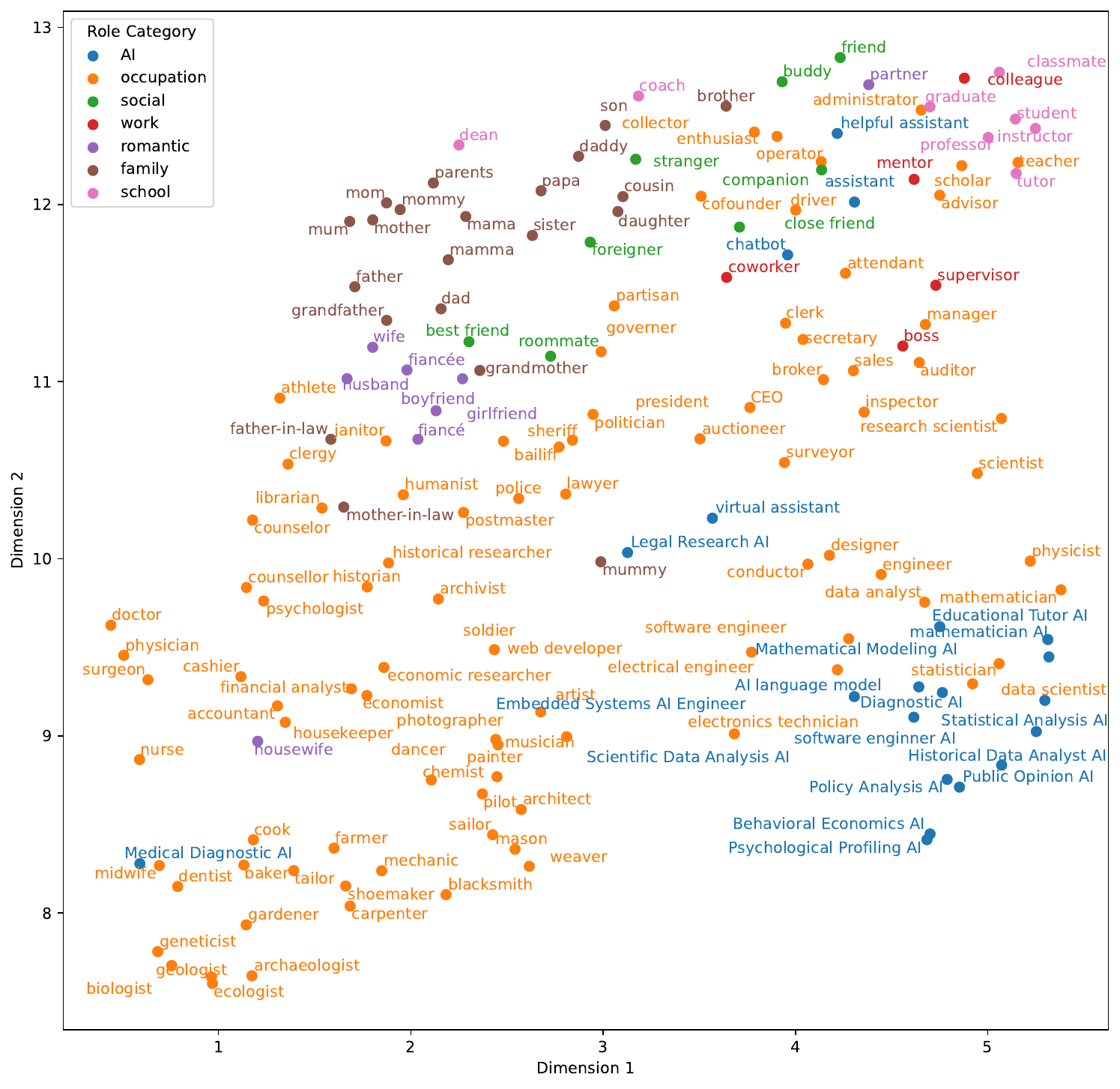}
    \caption{Role embeddings calculated by UMAP for Llama3-70B}
    \label{fig:role-embed-llama-70B}
\end{figure*}

\begin{figure*}[ht]
    \centering    \includegraphics[width=\textwidth]{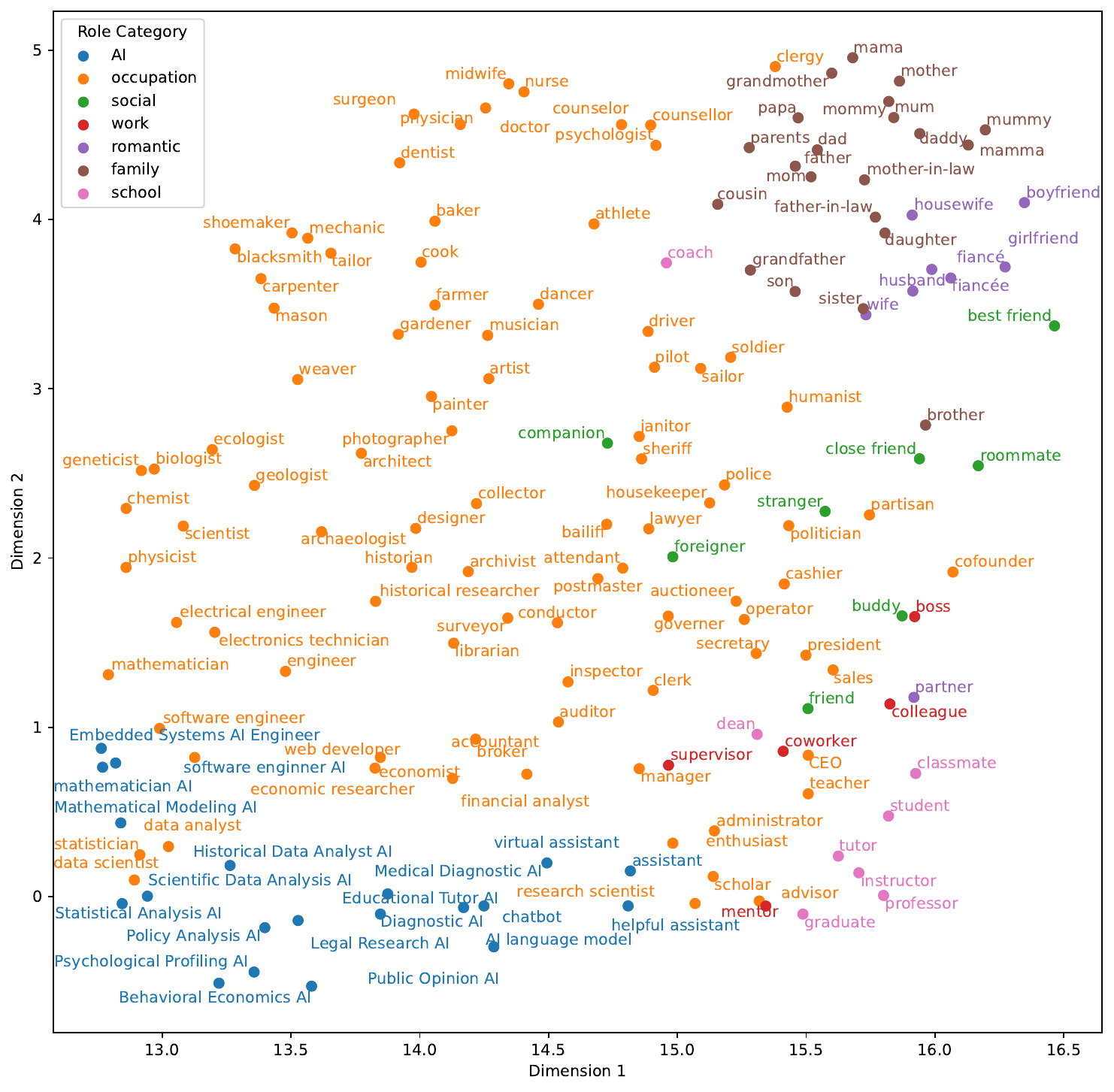}
    \caption{Role embeddings calculated by UMAP for Qwen2.5-7B}
    \label{fig:role-embed-qwen-7B}
\end{figure*}

\begin{figure*}[ht]
    \centering    \includegraphics[width=\textwidth]{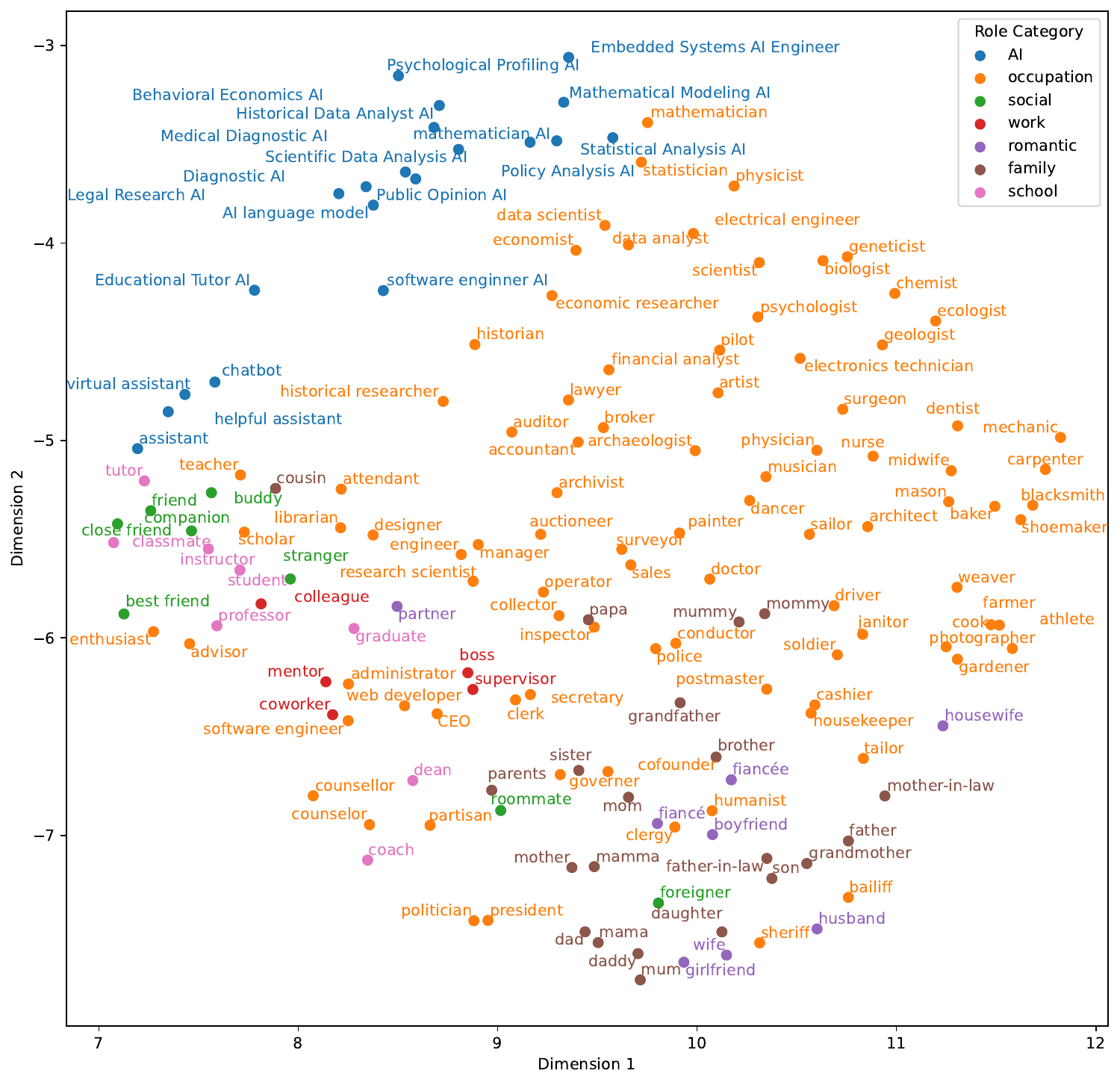}
    \caption{Role embeddings calculated by UMAP for Qwen2.5-72B}
    \label{fig:role-embed-qwen-72B}
\end{figure*}

\begin{figure*}[ht]
    \centering
    \begin{subfigure}{0.45\textwidth}
        \centering
        \includegraphics[width=\linewidth]{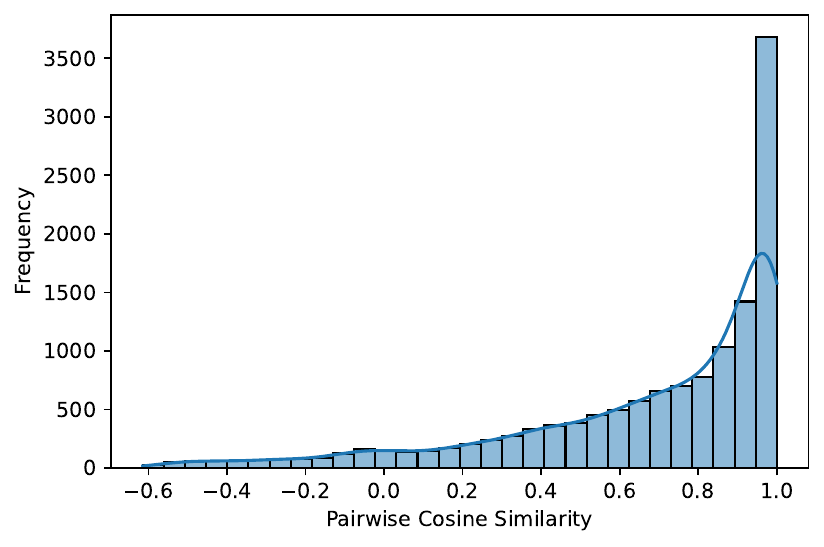}
        \caption{FLAN-T5-XXL}
        \label{fig:flan-role-cos}
    \end{subfigure}
    \begin{subfigure}{0.45\textwidth}
        \centering
        \includegraphics[width=\linewidth]{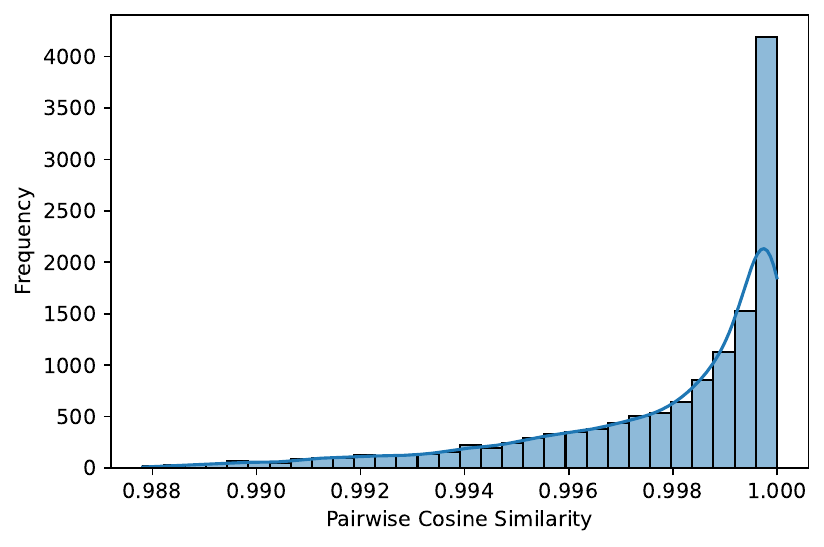}
        \caption{Mistral-7B}
        \label{fig:mistral-role-cos}
    \end{subfigure}
    \\
    \begin{subfigure}{0.45\textwidth}
        \centering
        \includegraphics[width=\linewidth]{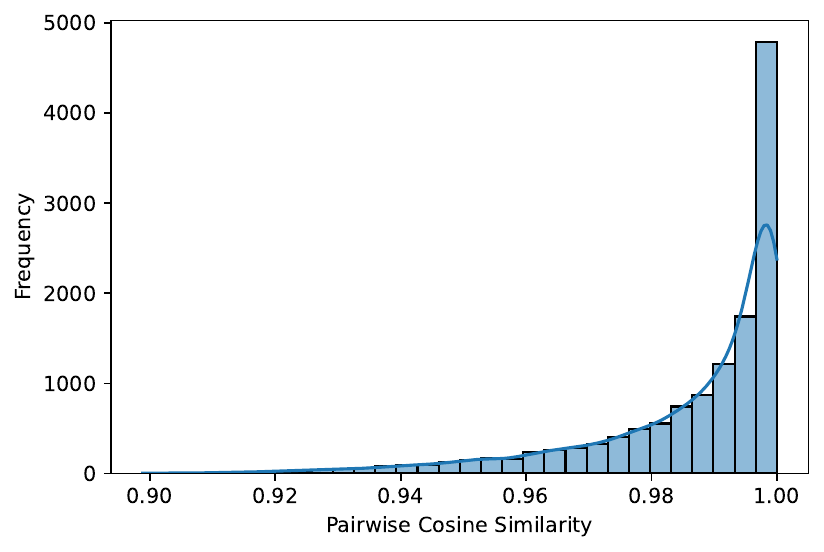}
        \caption{Llama3-8B}
        \label{fig:llama-8B-role-cos}
    \end{subfigure}
    \begin{subfigure}{0.45\textwidth}
        \centering
        \includegraphics[width=\linewidth]{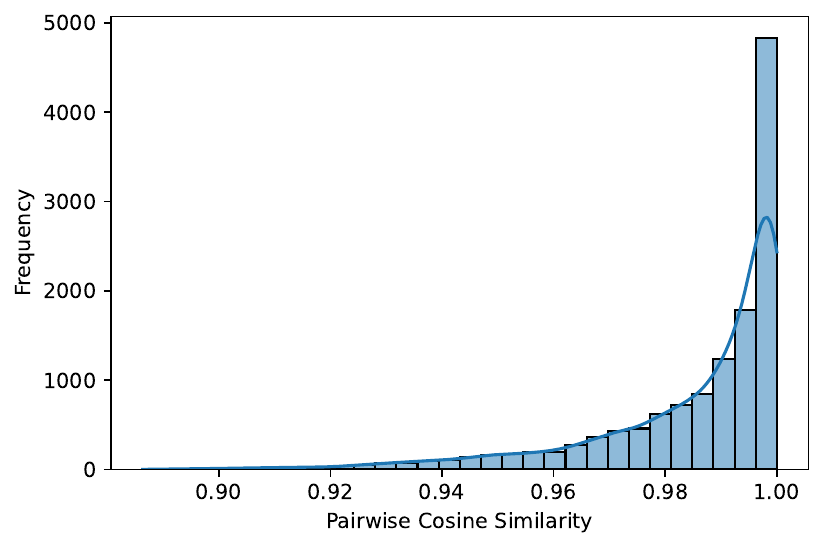}
        \caption{Llama3-70B}
        \label{fig:llama-70B-role-cos}
    \end{subfigure}
    \\
    \begin{subfigure}{0.45\textwidth}
        \centering
        \includegraphics[width=\linewidth]{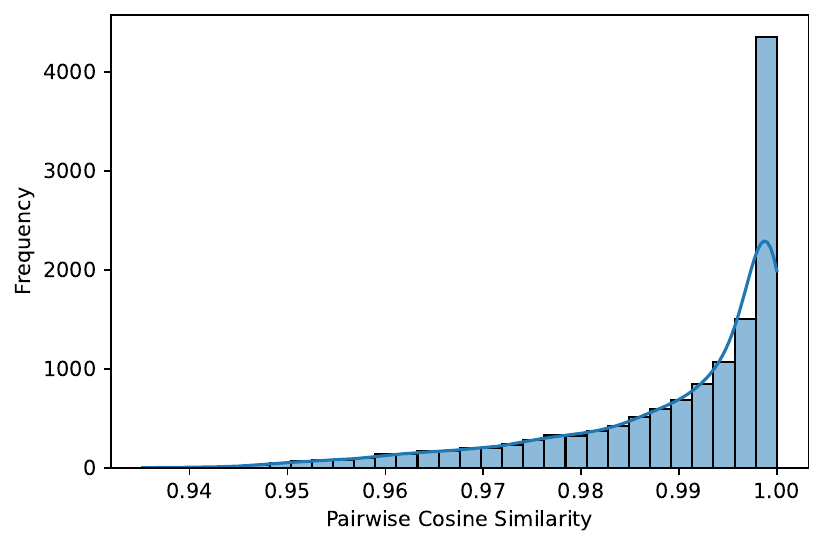}
        \caption{Qwen2.5-7B}
        \label{fig:qwen-7Brole-cos}
    \end{subfigure}
    \begin{subfigure}{0.45\textwidth}
        \centering
        \includegraphics[width=\linewidth]{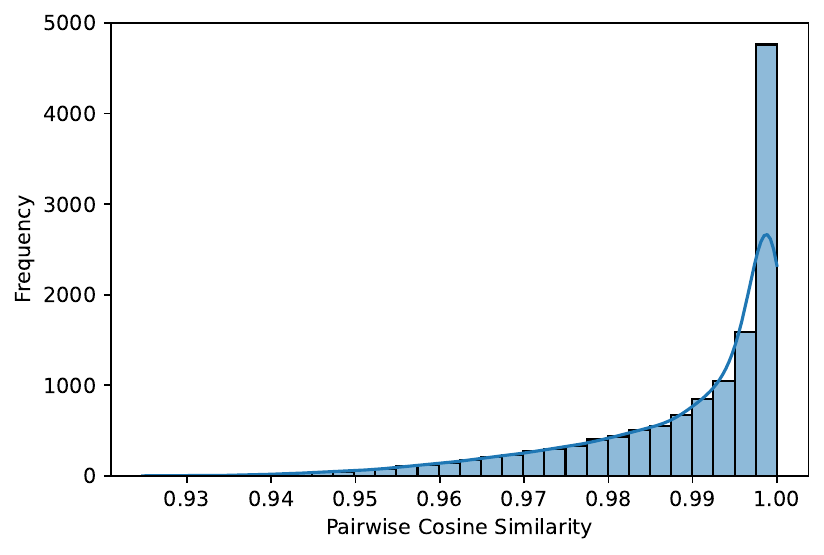}
        \caption{Qwen2.5-72B}
        \label{fig:qwen-72B-role-cos}
    \end{subfigure}
    \caption{Cosine similarity distribution of role embeddings for each model.}
    \label{fig:role-cos-similarity-all-models}
\end{figure*}

To quantify the performance differences of various personas, we build embeddings for each persona and analyze the similarity across these embeddings. For each persona, we first calculate the average accuracy of each question, resulting in a vector of length 2410. Then, we use Uniform Manifold Approximation and Projection (UMAP) for dimension reduction to map these embeddings to two dimensions. The persona embeddings calculated from each model are illustrated in Figure~\ref{fig:role-embed-flan},  \ref{fig:role-embed-mistral}, \ref{fig:role-embed-llama-8B}, \ref{fig:role-embed-llama-70B}, \ref{fig:role-embed-qwen-7B}, and Figure~\ref{fig:role-embed-qwen-72B}. The distributions of pairwise cosine similarity for each model are shown in Figure~\ref{fig:role-cos-similarity-all-models}. The skewed distributions in models Llama3, Mistral, and Qwen2.5 towards the right around value 1 demonstrate the high similarity across roles, whereas the embeddings are relatively more divergent in Flan-T5. 

\section{Model Consistency}

\begin{figure}[ht]
    \centering    
    \includegraphics[width=\columnwidth]{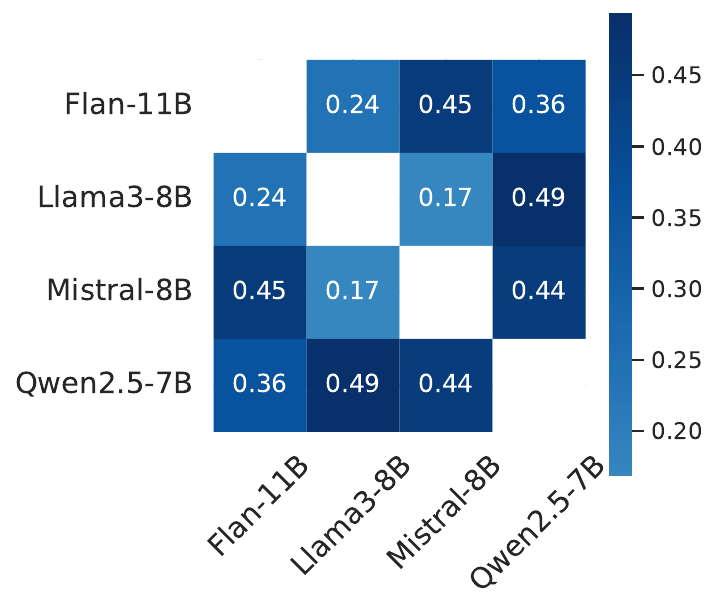}
    \caption{Heatmap of the correlation between personas' mean accuracy across middle-sized models.}
    \label{fig:model-consistency}
\end{figure}

The correlation between personas' mean accuracy over 2410 questions and 4 prompts across 4 middle-sized models are illustrated in Figure~\ref{fig:model-consistency}. 

\end{document}